\documentclass{article}
\usepackage{ijcai24}

\usepackage{times}
\usepackage{soul}
\usepackage{url}
\usepackage[hidelinks]{hyperref}
\usepackage[utf8]{inputenc}
\usepackage[small]{caption}
\usepackage{graphicx}
\usepackage{amsmath}
\usepackage{amsthm}
\usepackage{booktabs}
\usepackage{algorithm}
\usepackage{algorithmic}
\usepackage[switch]{lineno}
\usepackage{color}
\usepackage[table]{xcolor}
\usepackage{algorithm}
\usepackage{algorithmic}
\usepackage{bbding}
\usepackage{pifont}
\usepackage{wasysym}
\usepackage{amssymb}
\usepackage{enumitem,amssymb}
\newlist{todolist}{itemize}{2}
\setlist[todolist]{label=$\square$}
\usepackage{pifont}

\title{Explore Internal and External Similarity for Single Image Deraining with 
\\
Graph Neural Networks}

\author{
    PaperID: 334
}

\author{
Cong Wang$^{1,2}$
\and
Wei Wang$^1$\thanks{Wei Wang is the corresponding author} \and
Chengjin Yu$^{3}$\And
Jie Mu$^4$\\
\affiliations
$^1$Shenzhen Campus, Sun Yat-sen University\\
$^2$The Hong Kong Polytechnic University\\
$^3$Anhui University\\
$^4$Dongbei University of Finance and Economics\\
\emails
supercong94@gmail.com,
wangwei29@mail2.sysu.edu.cn,
23073@ahu.edu.cn,
jiemu@dufe.edu.cn
}

\begin{document}

\maketitle

\begin{abstract}
Patch-level non-local self-similarity is an important property of natural images. However, most existing methods do not consider this property into neural networks for image deraining, thus affecting recovery performance. Motivated by this property, we find that there exists significant patch recurrence property of a rainy image, that is, similar patches tend to recur many times in one image and its multi-scale images and external images. To better model this property for image detaining, we develop a multi-scale graph network with exemplars, called MSGNN, that contains two branches: 1) internal data-based supervised branch is used to model the internal relations of similar patches from the rainy image itself and its multi-scale images and 2) external data-participated unsupervised branch is used to model the external relations of the similar patches in the rainy image and exemplar. Specifically, we construct a graph model by searching the k-nearest neighboring patches from both the rainy images in a multi-scale framework and the exemplar. After obtaining the corresponding k neighboring patches from the multi-scale images and exemplar, we build a graph and aggregate them in an attentional manner so that the graph can provide more information from similar patches for image deraining. We embed the proposed graph in a deep neural network and train it in an end-to-end manner. Extensive experiments demonstrate that the proposed algorithm performs favorably against eight state-of-the-art methods on five public synthetic datasets and one real-world dataset. The source codes will be available at https://github.com/supersupercong/MSGNN.
\end{abstract}

\section{Introduction}
Images captured in a rainy environment are usually corrupted by rain streaks. These rainy images seriously degrade visibility, which accordingly interferes with the following high-level tasks, e.g., semantic segmentation, object detection, etc.
Therefore, it is of great need to develop an effective deraining algorithm to improve the quality of the images so that they can facilitate related applications.

Mathematically, the degradation model for image deraining is usually formulated as a linear combination of a rain streaks component $R$ with a clean background image $B$:
\begin{equation}
O = B + R.
\label{eq:rain model}
\end{equation}
According to (\ref{eq:rain model}), image deraining is a highly ill-posed problem as the rain streaks $R$ and the clear image $B$ are both unknown. 
To make this problem well-posed, conventional methods~\cite{derain_dsc_luo,derain_lowrank,derain_lp_li} usually impose priors on rain streaks and clear images to constrain the solution spaces. 
\begin{figure}[!t]
\begin{center}
\includegraphics[width = 1\linewidth]{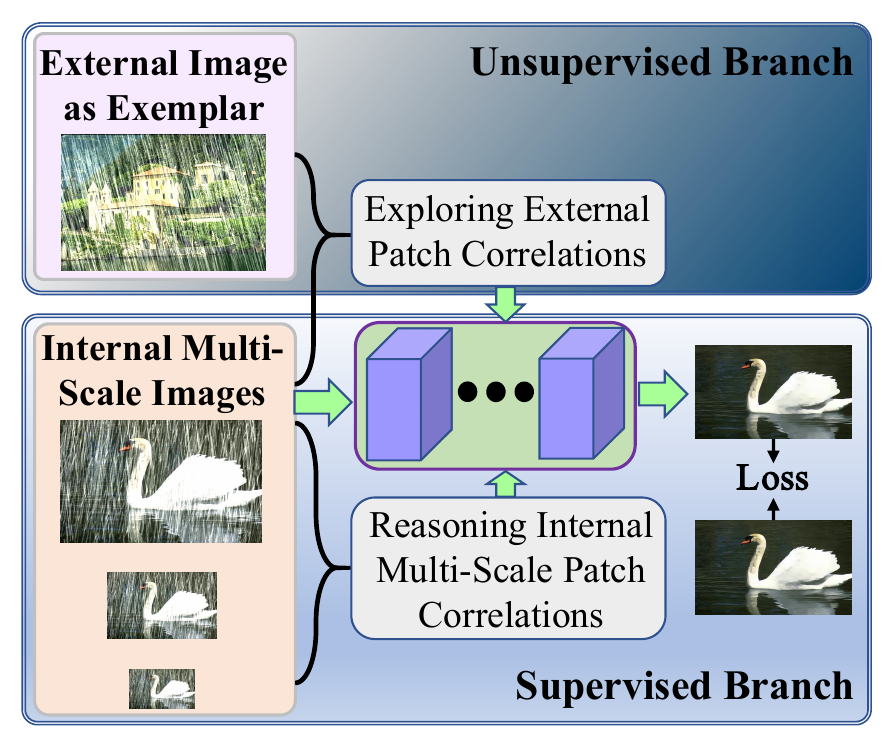}
\end{center}
\caption{Illustration of our main idea.
The proposed method consists of two branches: \textbf{Supervised Branch} and \textbf{Unsupervised Branch}.
The Supervised Branch reasons the internal patch correlation on the multi-scale images, while the unsupervised branch utilizes the learning ability of graph neural networks to explore the external patch correlation with unseen images as exemplars to learn more rainy conditions for better rain removal.
}
\label{fig: mianidea}
\end{figure}
However, these priors-based methods usually lead to complex optimization problems that are difficult to solve and have higher time consumption.

Inspired by the success of the deep convolutional neural networks (CNNs) in many vision tasks, recent efforts have been devoted to applying CNNs to solve the image restoration~\cite{jin2022structure,jin2023enhancing,jin2022unsupervised,wang2021single,wang2022online,wang2024powerful,wang2024promptrestorer,wang2024selfpromer,wang2024uhdformer,cui2022semi,zhu2020physical,xu2019segmentation,xu2022dnas,xu2023towards} and also deraining problem~\cite{derain_jorder_yang,derain_rescan_li,derain_zhang_did,Wang_2019_ICCV,derain_2019_CVPR_spa,Derain-cvpr19-semi,derain_prenet_Ren_2019_CVPR,mm20_wang_jdnet,cvpr20_syn2real,cvpr20_wang_rcdnet,derain_mprnet_cvpr21,Zhou_2021_CVPR}.
These deraining methods generally model the image deraining problem as a pixel-wise image regression process that directly learns the mapping from rain images to clear images by end-to-end trainable networks.
Although these CNN-based approaches have achieved decent results, few of them consider patch recurrence into neural networks and hence are less effective in modeling the non-local self-similarity property of natural images. Although using large receptive fields can remedy this problem to some extent, it usually leads to larger or deeper networks that are difficult to train.

To better solve the image deraining task, we propose an effective method that explores the non-local similarity property of rainy images by a graph neural network. 
Figure~\ref{fig: mianidea} illustrates the main idea of our method that contains two branches: 1) internal data-based supervised branch is used to model the internal relations of similar patches from the rainy image itself and its multi-scale images and 2) external data-participated unsupervised branch is used to model the external relations of the similar patches in the rainy image and exemplar.
\begin{figure}[!t]
\begin{center}
\begin{tabular}{c}
\hspace{-2.5mm}\includegraphics[width = 1.014\linewidth]{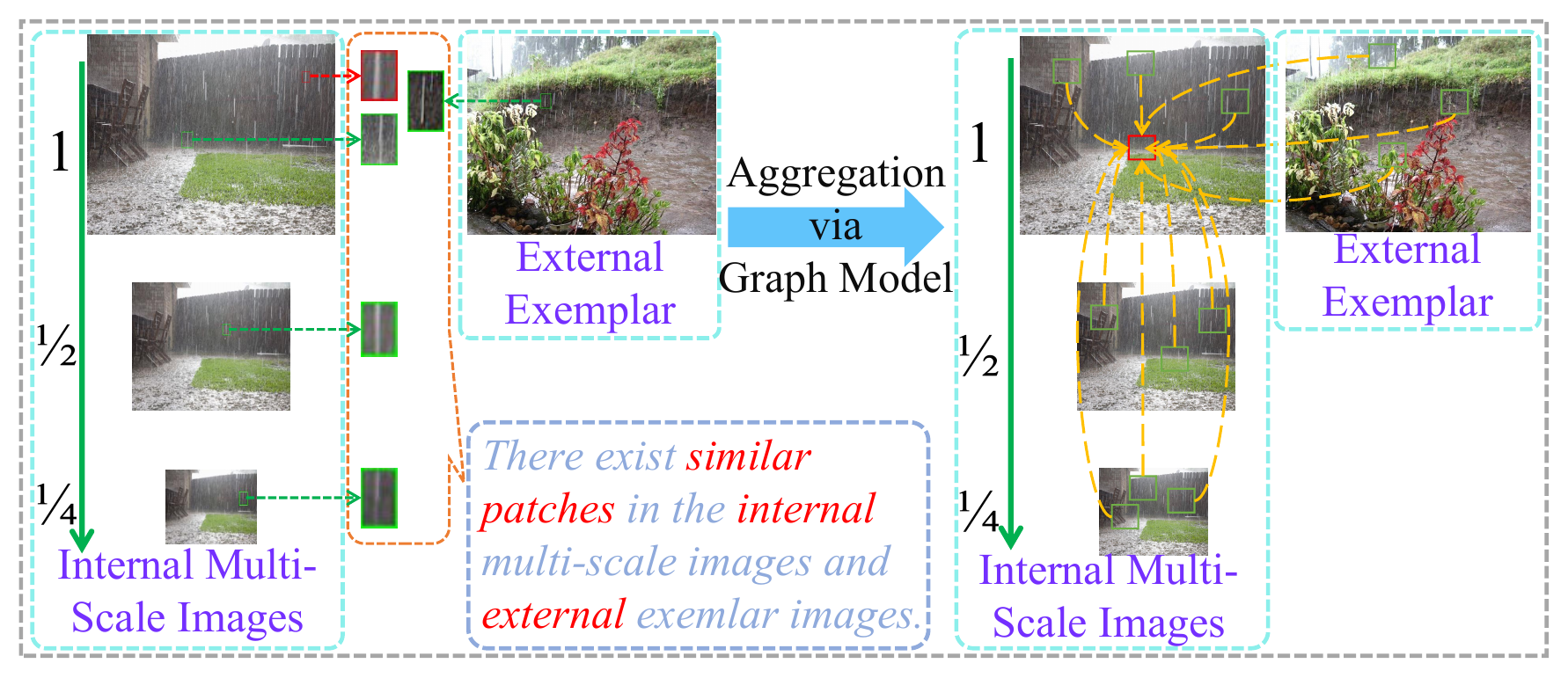}
\end{tabular}
\end{center}
\caption{Illustration of the internal similarity in multi-scale images and external similarity in the exemplar.
As can be seen, patch recurrence property indeed exists in internal multi-scale images and external exemplars.
}
\label{fig: multiscaleimage}
\end{figure}

The proposed method is motivated by the patch recurrence property of natural images. We find that similar patches tend to recur many times in one image and its multi-scaled versions and external images, especially in rainy images as shown in Figure~\ref{fig: multiscaleimage}, where most image patches in rainy images have similar rain streaks. This property stimulates us to build the non-local graph modules that can aggregate these neighbors that have similar rain streaks into one query patch to achieve better deraining.
To model the non-local similarity property of both the rainy image itself and external exemplars, we first model the internal relations of similar patches from the rainy image itself by a multi-scale graph network. Then, we explore the property of the exemplars and model the external relations of the similar patches in the rainy images and exemplars. 
Specifically, we construct a graph model by searching $k$-nearest neighboring patches in the multi-scale images and exemplar for each query patch of the input rainy image. 
We then obtain the corresponding $k$ neighboring patches in the multi-scale images and exemplar and aggregate them with a graph attention mechanism. 
Finally, we formulate the constructed graph into an end-to-end trainable deep neural network to solve image deraining. 
The main contributions of this paper are summarized as follows:
\begin{itemize}
\item We propose a multi-scale graph module to explore internal non-local similarity by aggregating similar patches in multi-scale rainy images to the query patch of the input rainy image.

\item We propose to use an exemplar image to explore external non-local similarity for enriching the representation of the query patch to furthermore improve the deraining quality.

\item We analyze the proposed algorithm and show that it is able to remove rain streaks and preserve image details. Quantitative and qualitative experiments demonstrate that the proposed algorithm outperforms state-of-the-art methods on both synthetic and real-world datasets.
\end{itemize}

\textit{\textbf{To the best of our knowledge, this is the first algorithm that explores the non-local property from both the rainy image itself and external exemplars by a graph neural network for single image deraining.}}
\section{Related Work}
\subsection{Single Image Deraining}

\noindent \textbf{Priors-based deraining methods} are typically built on some assumption about rain streaks and rainy images~\cite{derain_dsc_luo,derain_lowrank,derain_lp_li}.
\begin{figure*}[!t]
\begin{center}
\includegraphics[width = 1\linewidth]{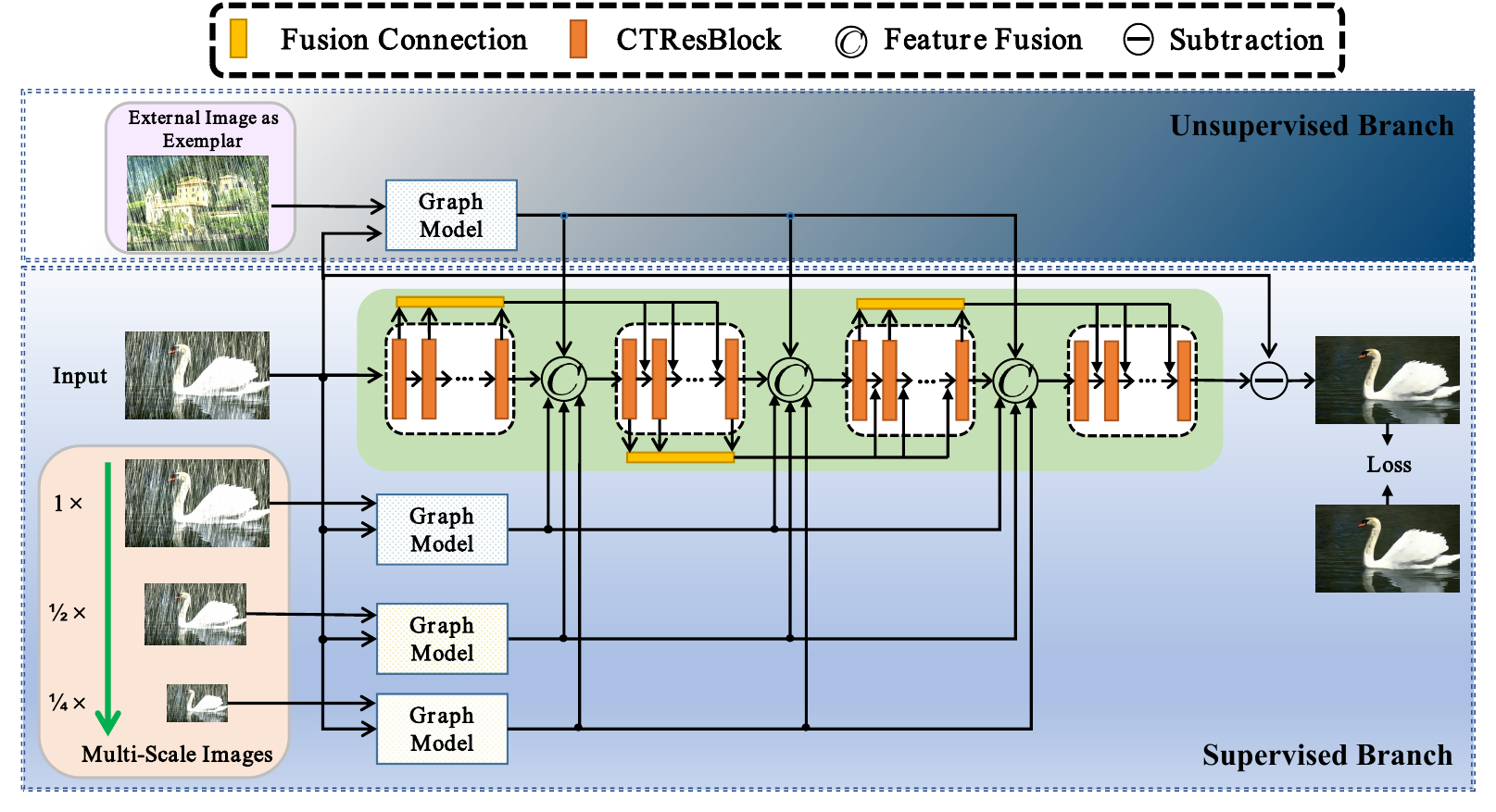}
\end{center}
\caption{Diagram of the proposed MSGNN. Our method consists of two branches: an internal data-based supervised branch and an external data-participated unsupervised branch. In the supervised branch, we explore the internal property of multi-scale images by reasoning the multi-scale patch correlations to help the original-scale image learn more internal image information. In the unsupervised branch, we build the external patch correlations between input images and another external image as an exemplar by utilizing the superiority of the graph network which can search similar patches between any two samples so that the network is able to learn more rainy conditions for better rain removal.
Graph Model is illustrated in Figure~\ref{fig: graph model}.
}
\label{fig: Overall Framework}
\end{figure*}

\noindent \textbf{CNNs-based deraining approaches}
have dominated recent research and achieved great success.
\cite{derain_ddn_fu} maps the high-frequency structure of a rainy image to negative residual rain streaks and obtains final rain-free images by \eqref{eq:rain model}.
After that, a series of deraining methods are proposed~\cite{derain_rescan_li,derain_2019_CVPR_spa,derain_prenet_Ren_2019_CVPR,mm20_wang_dcsfn,mm20_wang_jdnet,wang2022online}
\subsection{Non-local Self-similarity for Image Restoration}
The non-local self-similarity prior assumes that similar patches frequently recur in a natural image, which has been exploited by many classical image restoration methods~\cite{Denoising_Transform-Domain,non-local-denoising,Neighbor_Embedding_SR}. 
Here, we briefly review recent approaches that exploit this prior with graph neural networks.
\citeauthor{n3net-nips}~\cite{n3net-nips} are the first to introduce a graph-based model for image denoising and image super-resolution.
They propose a neural nearest neighbor network that learns to aggregate the neighbors of the query patch.
\citeauthor{ignn-zhou-nips}~\cite{ignn-zhou-nips} propose a cross-scale graph network for image super-resolution.

In this paper, we explore the non-local similarity property of rainy images by taking into consideration of multi-scale input images and external exemplars.

\section{Proposed Method}

\subsection{Overall Framework}\label{sec:Overall Framework}
The overall framework is shown in Figure~\ref{fig: Overall Framework}. Our method intends to learn the rain streaks from an input image via a graph-boosted CNN, and then subtract the learned rain streaks from the input image to obtain the deraining result.
We express this process as follows:
\begin{equation}\small
\begin{array}{ll}
\tilde{B} =\text{MSGNN}\Big(O,\mathcal{R}(O,O),
\mathcal{R}(O,O_{\frac{1}{2}}),\mathcal{R}(O,O_{\frac{1}{4}}), \mathcal{R}(O,E)\Big),
\end{array}
\label{eq:msgnn}
\end{equation}
where $\text{MSGNN}$ denotes our proposed backbone CNN.
$\tilde{B}$ refers to the estimated rain-free image.
$O_{\frac{1}{2}}$ and $O_{\frac{1}{4}}$ refers to the $\frac{1}{2}$ and $\frac{1}{4}$ scaled images, respectively, while $E$ represents the external exemplar.
$\mathcal{R}(\cdot,\cdot)$ denotes a graph model that is divided into two steps: the nearest neighbor search and attentional aggregation.

$\text{MSGNN}$ takes a rainy image $O$ as input and reasons internal multi-scale patch correlation with its multi-scale images and explores the external patch correlation with an exemplar in the network to search the most similar patch information and aggregate them for better rain streaks removal.

The $\text{MSGNN}$ contains $N$ sub-networks with an identical structure, and each sub-network consists of $M$ Channel Transformation~\cite{ct_yang_cvpr20} Residual Blocks (CTResBolcks).
Two adjacent sub-networks are connected via ``fusion connection'' (yellow blocks), which concatenates all the outputs of the convolution layers, then applies a $1 \times 1$ convolution on them, and sends the results to the next sub-network.
The output of the graph model is injected into every direct connection between sub-networks via feature fusion, which concatenates all features together and then applies two convolution layers with kernel size $5 \times 5$ and stride size $2 \times 2$.
The graph model is described later in the rest of this section, and the details of CTResBolck are provided in supplementary material.

To train the network, we minimize the negative structural similarity index measure (SSIM)~\cite{SSIM_wang} between the estimated rain-free image $\tilde{B}$ and the ground truth $B$:
\begin{equation}
\mathcal{L} = -\text{SSIM}(\tilde{B}, B).
\label{eq:ssim-loss}
\end{equation}

\subsection{Graph Model}\label{sec:Graph Module}
Our graph model exploits both internal and external non-local similarity. Internal non-local similarity is utilized to identify similar patches to the query patch within the input image of different scales. The external non-local similarity is to find similar patches from another rainy image, the external exemplar. 

\begin{figure}
\begin{center}
\begin{tabular}{c}
\hspace{-2mm}\includegraphics[width = 1.014\linewidth]{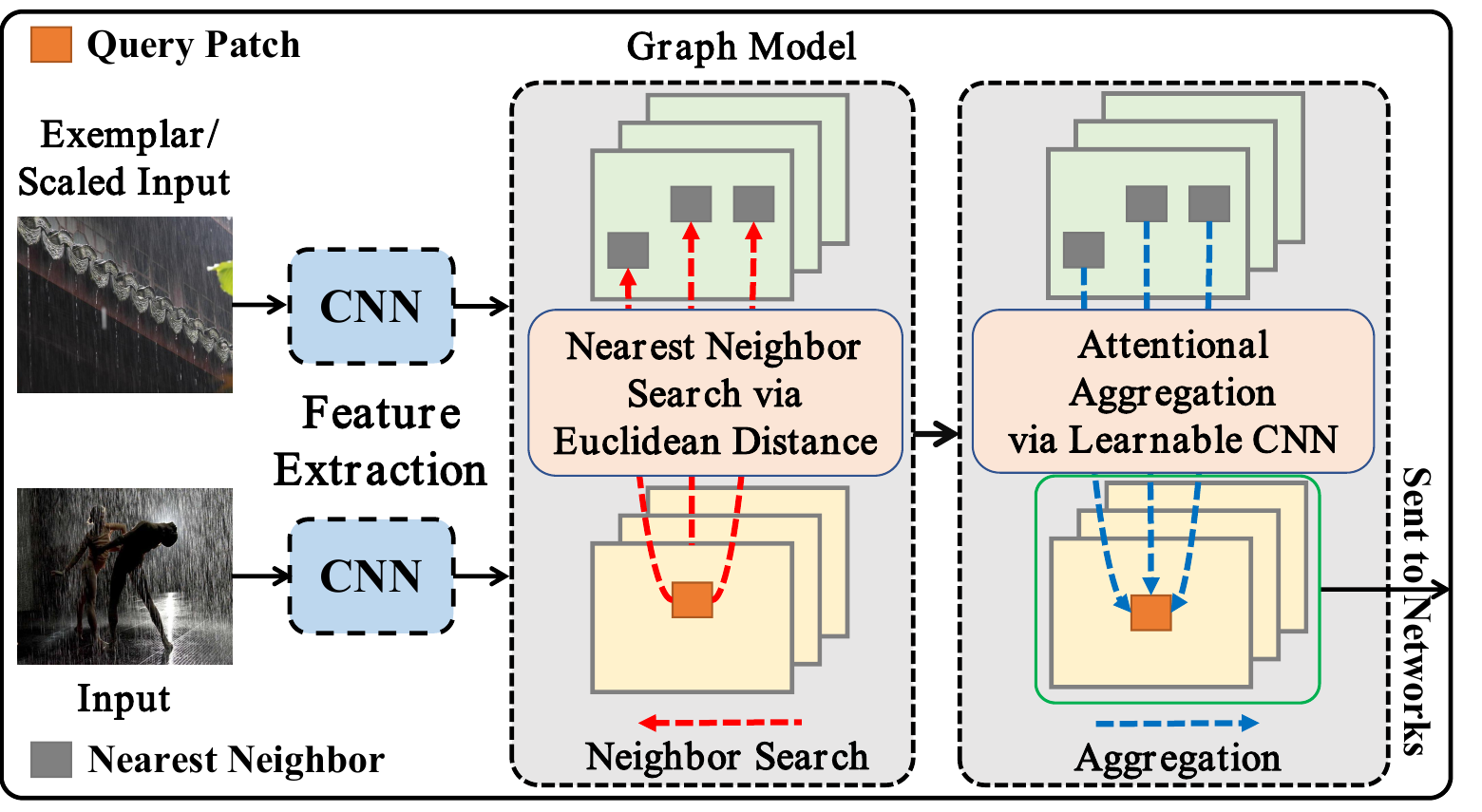}
\end{tabular}
\end{center}
\caption{Illustration of the graph model which is divided into two steps: nearest neighbor search and attentional aggregation.
}
\label{fig: graph model}
\end{figure}

As shown in Figure~\ref{fig: graph model}, the proposed graph model consists of the following steps: 1) extract features from images by a CNN; 2) for each query patch, search for patches with similar feature maps; 3) aggregate the similar patches with attention mechanism and then send the combined features to the backbone CNN to facilitate the deraining of the query patch.

\subsubsection{Internal and External Similarity}\label{sec:similarity}
As mentioned above, non-local self-similarity has been exploited in previous graph-based methods for various restoration tasks~\cite{n3net-nips,ignn-zhou-nips}.
However, these methods overlook information in multi-scale images.
In this paper, we explore the non-local self-similarity property in multi-scale inputs as in \cite{cvpr20_jiang_mspfn}.
Specifically, patch recurrence not only exists in the original image but also can be found across its different scaled versions due to the perspective nature of images -- recurrent patterns may be located at different distances and thus have different visual sizes and are similar to each other only after properly scaled.

In addition to internal self-similarity, external images of the same domain, topic, or theme can also be similar to the input image and thus can provide more information for deraining. These external images are referred to as exemplars.
Hence, in our model, we also explore the relation between the input image and an exemplar image for better deraining.
Figure~\ref{fig: multiscaleimage} provides a better illustration of the internal patch similarity in multi-scale images and external patch similarity in exemplars, which supports our motivation.

\subsubsection{Nearest Neighbor Search}\label{sec:Nearest Neighbor Search }

For a query patch of the input image, we want to find similar patches from each scaled input image and the exemplar image. Here, we take the exemplar image as an example to illustrate the graph construction process, which can be similarly applied to other images.
The graph construction process includes the following steps: we first extract features from the original rainy image and the exemplar image via a three-layer CNN, and denote them by $\mathcal{F}_{O}$ and $\mathcal{F}_{E}$ respectively. Then, we divide $\mathcal{F}_{O}$ and $\mathcal{F}_{E}$ into patches of size $l \times l$. 
For each query patch in $F_{O}$, which is indexed by $q$ and denoted by $\mathcal{P}_{O}^{q}$, we find $k$ nearest patches $\mathcal{P}_{E}^{n_{r}}$ among patches in $\mathcal{F}_{E}$, based on the Euclidean distance between their CNN-extracted features. Nearest patches $\mathcal{P}_{E}^{n_{r}}$ are indexed by $n_r$ and $r = \{1, \cdots, k\}$.

As such, a $k$-nearest neighbor graph $\mathcal{G}_{k}(\mathcal{V}, \mathcal{E})$ is constructed. 
$\mathcal{V}$ is the patch set (vertices in the graph) and equal to $\mathcal{V}_O \cup \mathcal{V}_E$, the union of input rainy image patch set $\mathcal{V}_{O}$ and an exemplar patch set $\mathcal{V}_{E}$. 
Set $\mathcal{E}$ is the edge set with size
$|\mathcal{E}| = |\mathcal{V}_{O}| \times k$, which indicates $k$ nearest neighbors for each query patch in $\mathcal{V}_{O}$. For each edge, one of its two terminal vertices is the input rainy image patch and the other one is the exemplar patch. 
\subsubsection{Attentional Aggregation}\label{sec: Attentional Aggregation}
With the $k$-nearest neighbor graph, we can perform attentional aggregation to combine the nearest neighbors of the query patch. Notice that the query patch itself is its closest neighbor. Specifically, the aggregation is done by weighted averaging:
\begin{equation}
\widehat{\mathcal{P}}_{O}^{q} = \frac{1}{\delta_{q}} \sum_{n_{r}\in S_{q}} \alpha_{n_r\to q} \mathcal{P}_{E}^{n_{r}}, 
\label{eq:aggregation}
\end{equation}
where $\alpha_{n_r\to q}$ is the weight, $S_q$ is the set of the $k$ nearest neighbors of $q$, $\delta_{q} = \sum_{n_{r}\in S_{q}} \alpha_{n_r\to q}$ is the normalizing factor, and $\widehat{\mathcal{P}}_{O}^{q}$ is the aggregated patch representation.

The weight $\alpha_{n_r\to q}$ is computed by an attention mechanism, which is devised to measure the similarity between the query patch and its neighbors in the sense that similar neighbors should have a large weight. It is implemented as follows. First, we calculate the difference between the query and the neighbor:
\begin{equation}
    \mathcal{D}^{n_{r} \to q} = \mathcal{P}_{O}^{q} - \mathcal{P}_{E}^{n_{r}}. 
\end{equation}
Then, we obtain the attention weight by a CNN with trainable parameters $\theta$:
\begin{equation}
    \alpha_{n_r\to q} = \exp(\text{CNN}_\theta(\mathcal{D}^{n_{r} \to q})),
\end{equation}
where the exponential function $\exp(\cdot)$ ensures all weights are positive. Finally, for all query patches,  we assemble the aggregated patch representations $\widehat{\mathcal{P}}_{O}^{q}$ into $\widehat{\mathcal{F}}_{O}$ via $\mathtt{patch2img}$~\cite{n3net-nips} and send it to downstream networks.
\begin{table*}[!t]
\setlength{\tabcolsep}{1mm}
\centering
\scalebox{0.78}{
\begin{tabular}{lcccccccccccccccccccccc}
\toprule
& \multicolumn{2}{c}{RESCAN} 
&\multicolumn{2}{c}{PreNet}  & \multicolumn{2}{c}{SpaNet}& \multicolumn{2}{c}{DCSFN} & \multicolumn{2}{c}{MSPFN}& \multicolumn{2}{c}{RCDNet}

& \multicolumn{2}{c}{DualGCN}
& \multicolumn{2}{c}{MOSS}
& \multicolumn{2}{c}{MSGNN}    
\\
\midrule
Dataset      & PSNR&SSIM & PSNR&SSIM  & PSNR & SSIM& PSNR & SSIM& PSNR & SSIM& PSNR & SSIM& PSNR & SSIM & PSNR & SSIM& PSNR & SSIM      \\
\midrule
Rain200H 
& 26.661&0.8419  
& 27.640&0.8872  
&25.484 & 0.8584
&28.469&  0.9016
&25.553& 0.8039
&28.698& 0.8904  
&\textit{28.758}& \textit{0.9026}
&25.283& 0.8418

&\textbf{29.627}&\textbf{0.9178} \\
Rain200L
&36.993&0.9788
&36.487&0.9792   
& 36.075 &0.9774 
&37.847&  \textit{0.9842} 
& 30.367& 0.9219
& 38.400  &0.9841
&\textit{38.415} &0.9818
&32.138&0.9563

&\textbf{39.088}&\textbf{0.9869} \\
Rain1200      
&32.127&0.9028 
&27.307&0.8712  
& 27.099& 0.8082
&\textit{32.275}& \textit{ 0.9228} 
&30.382&0.8860
&32.273& 0.9111

&32.033& 0.9163
&31.668&0.9122
&\textbf{33.110}  &\textbf{0.9274} \\
Rain1400   
&30.969&0.9117 
&30.608&0.9181 
& 29.000& 0.8891
&\textit{31.493}&   \textit{0.9279}
&31.514& 0.9203
& 31.016  &0.9164
&30.567& 0.9148
&30.355&0.9177

&\textbf{31.703}& \textbf{0.9299} \\
Rain12     
&32.965& 0.9545
&34.791&0.9644 
&33.217&  0.9546
&35.803&   0.9683 
&34.253& 0.9469 
& 31.038  &0.9069

&\textit{35.805} &\textit{0.9687}
&30.775& 0.9317
&\textbf{36.799}& \textbf{0.9707} \\
\midrule
Parameter  
&\multicolumn{2}{c}{0.15M}  
&\multicolumn{2}{c}{0.17M}  
&\multicolumn{2}{c}{0.28M}  
&\multicolumn{2}{c}{6.45M}  
&\multicolumn{2}{c}{21.00M} 
&\multicolumn{2}{c}{3.67M}

& \multicolumn{2}{c}{2.73M}
&\multicolumn{2}{c}{2.86M}
& \multicolumn{2}{c}{2.46M}\\
\bottomrule
\end{tabular}}
\caption{Quantitative results on five synthetic datasets.
The best and second best results are marked in \textbf{bold} and \textit{italic} respectively.}
\label{tab: the results in synthetic datasets}
\end{table*}
\begin{figure*}[!t]
\begin{center}
\begin{tabular}{cccccccccc}
\includegraphics[width = 0.195\linewidth]{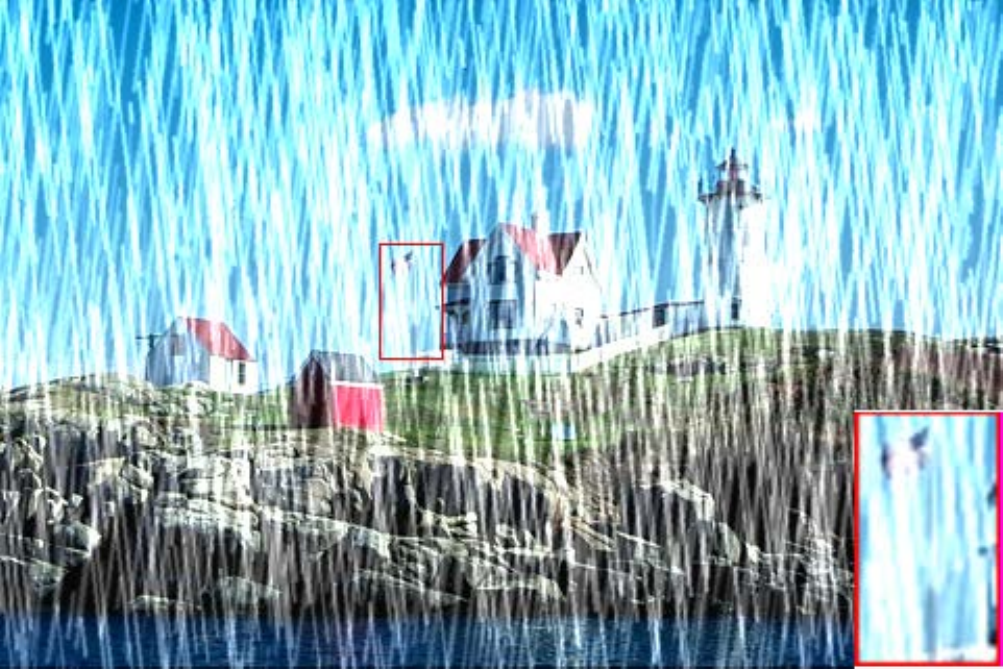} &\hspace{-4.5mm}
\includegraphics[width = 0.195\linewidth]{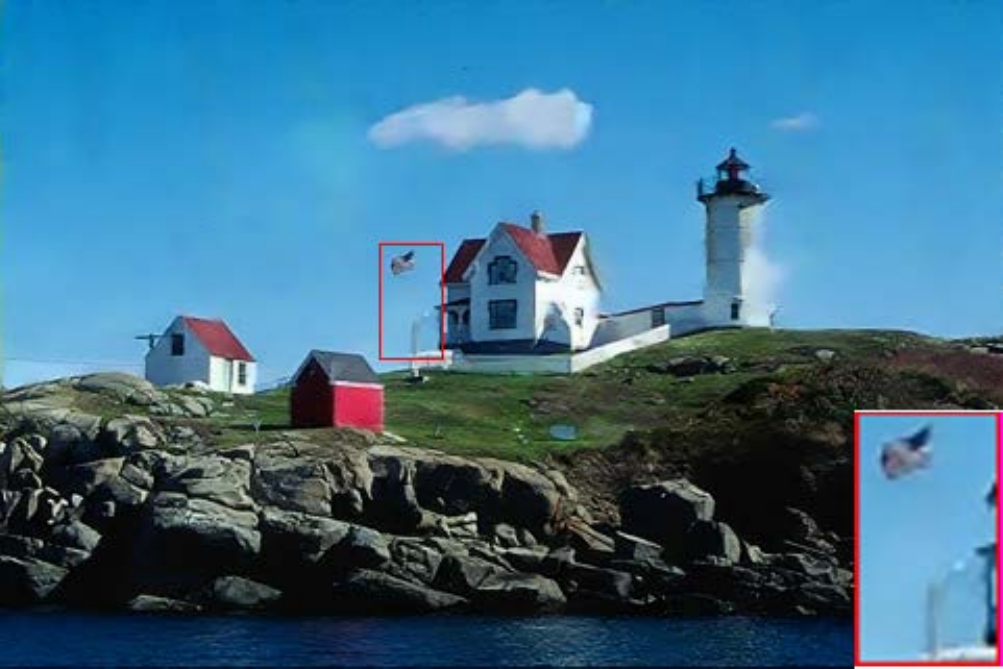} &\hspace{-4.5mm}
\includegraphics[width = 0.195\linewidth]{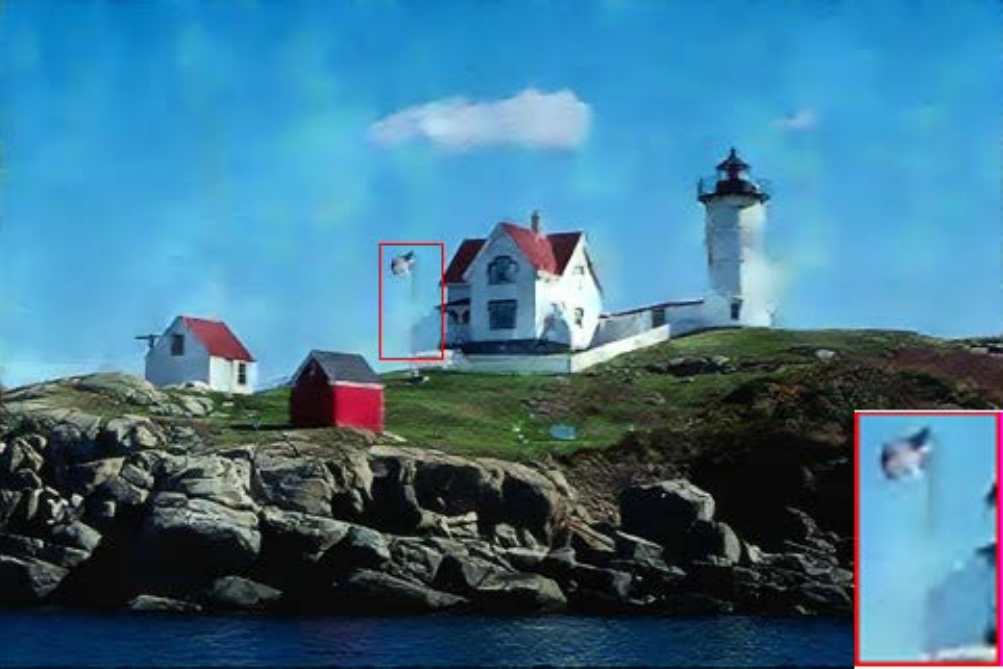} &\hspace{-4.5mm}
\includegraphics[width = 0.195\linewidth]{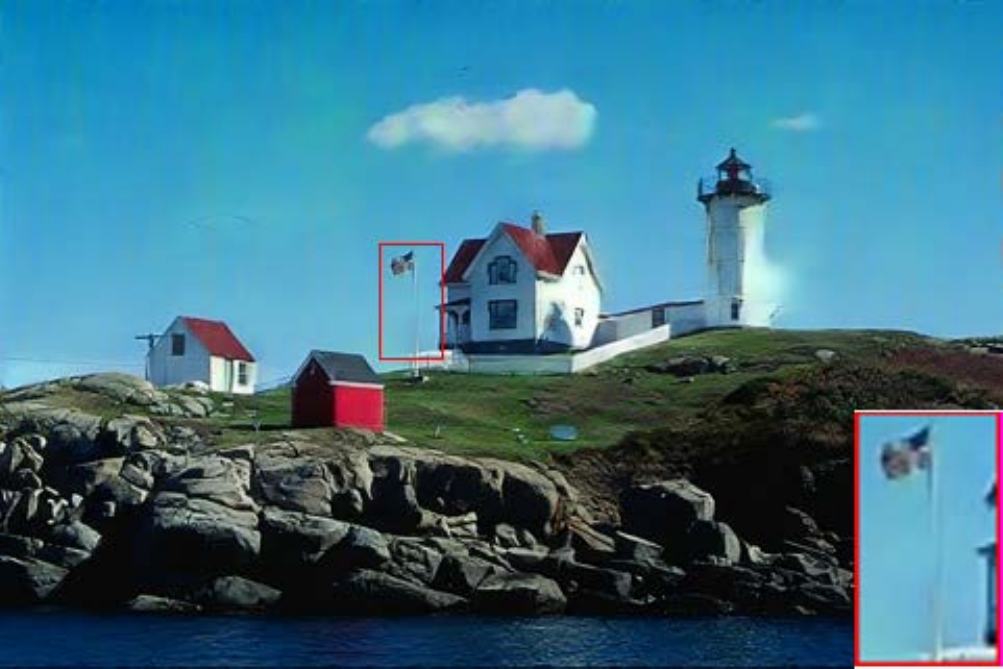} &\hspace{-4.5mm}
\includegraphics[width = 0.195\linewidth]{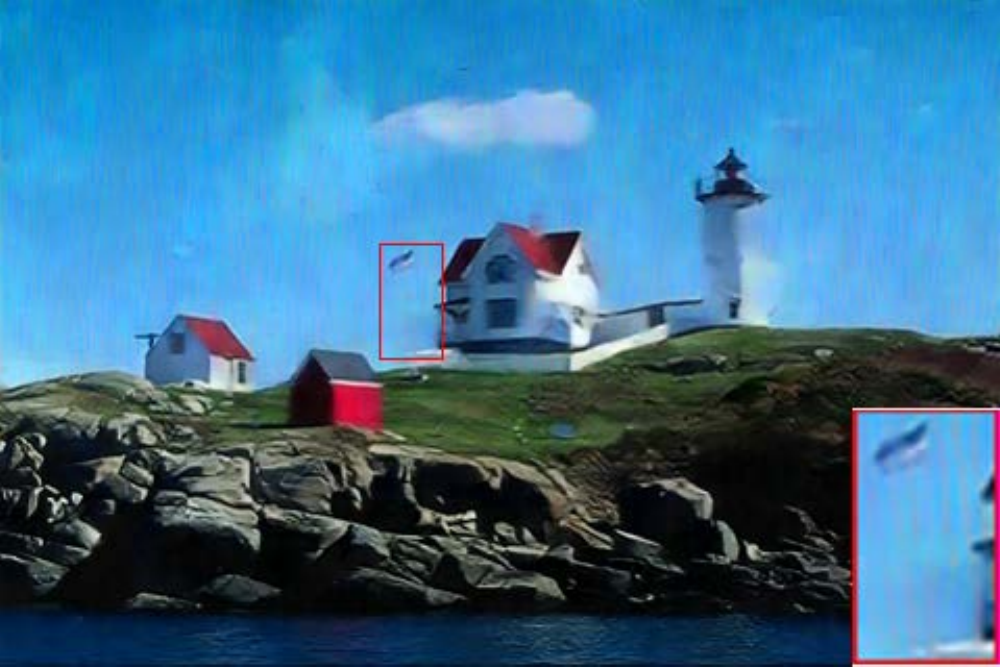} 
\\
(a) Input  &\hspace{-4.5mm}  (b) PreNet&\hspace{-4.5mm} (c) SpaNet&\hspace{-4.5mm} (d) DCSFN&\hspace{-4.5mm} (e) MSPFN
\\

\includegraphics[width = 0.195\linewidth]{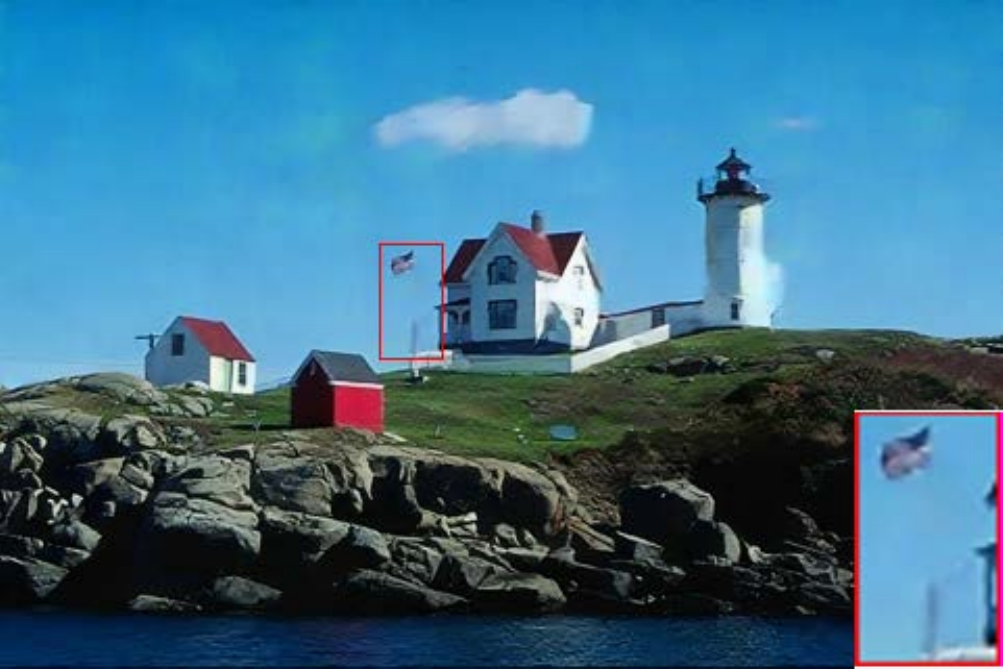} &\hspace{-4.5mm}
\includegraphics[width = 0.195\linewidth]{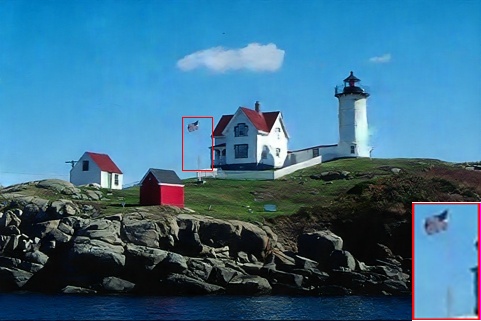} &\hspace{-4.5mm}
\includegraphics[width = 0.195\linewidth]{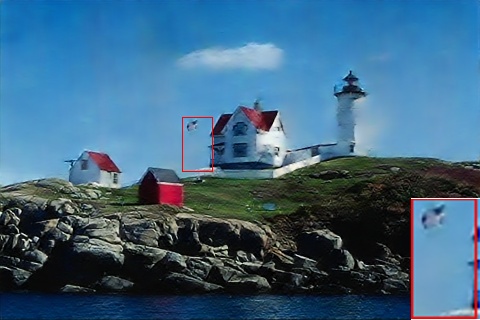} &\hspace{-4.5mm}
\includegraphics[width = 0.195\linewidth]{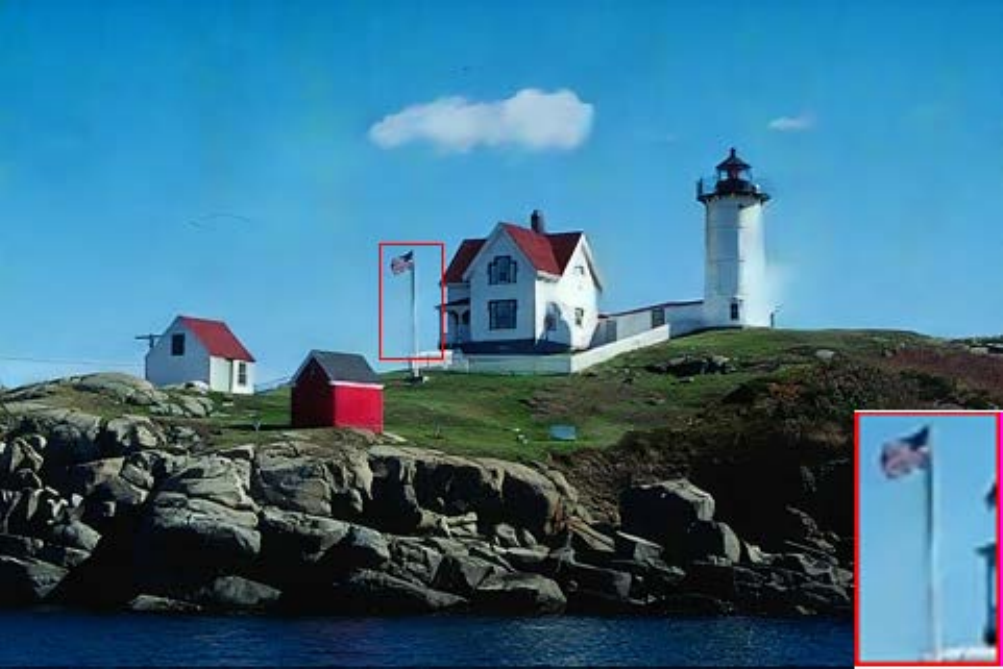} &\hspace{-4.5mm}
\includegraphics[width = 0.195\linewidth]{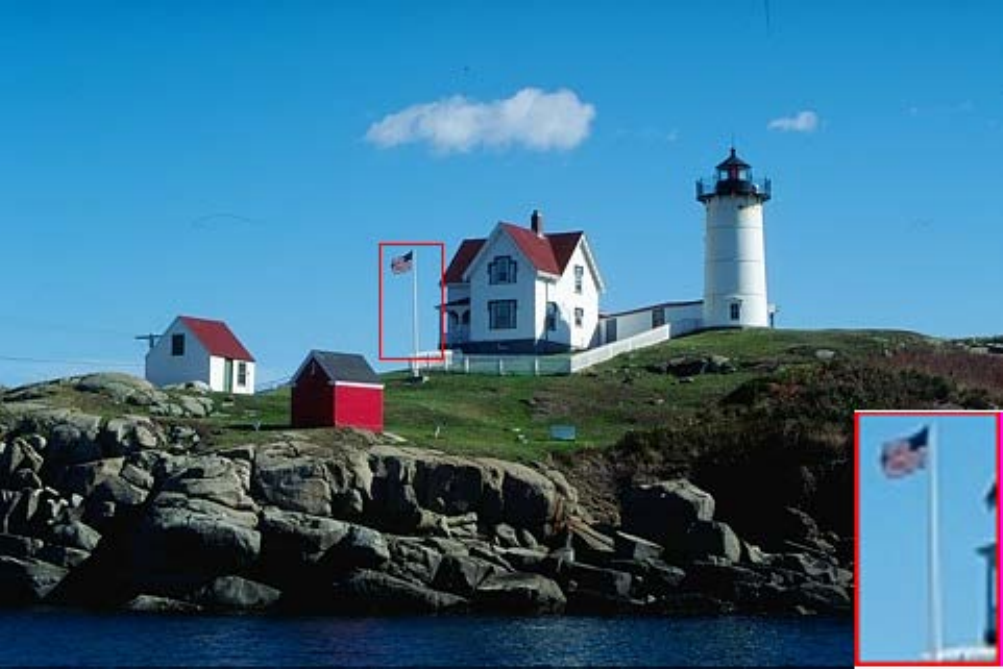} 
\\
(f) RCDNet&\hspace{-4.5mm} (g) DualGCN&\hspace{-4.5mm}(h) MOSS &\hspace{-4.5mm}(i) \textbf{MSGNN}&\hspace{-4.5mm} (j) GT
\end{tabular}
\end{center}
\caption{Comparisons with state-of-the-art methods on synthetic dataset.
Our proposed MSGNN is able to restore better texture.
}
\label{fig:deraining-syn-example}
\end{figure*}
\section{Experimental Results}
To demonstrate the effectiveness of our proposed method for single image deraining, we evaluate it on both synthetic datasets and real-world datasets.
For a fair comparison, we ensure that all the baselines are retrained using the codes provided by the authors.

\subsection{Datasets}
{\flushleft \bf Synthetic datasets.}
We conduct deraining experiments on five widely used synthetic datasets: Rain200H~\cite{derain_jorder_yang}, 
Rain200L~\cite{derain_jorder_yang}, 
Rain1200~\cite{derain_zhang_did},
Rain1400~\cite{derain_ddn_fu},
Rain12~\cite{derain_lp_li}.
Note that we use the model trained on Rain200H to test on the Rain12 dataset since Rain12 does not have training images.
We use the Rain100H as the analysis dataset.

{\flushleft \bf Real-world dataset.}
Note that \cite{derain_jorder_yang,deraining-benchmark-analysis,mm20_wang_jdnet} provide a mass of real-world rainy images. We use them as real-world datasets.
\subsection{Implementing Details}
We randomly select a rainy image as an exemplar from the dataset as input to the network.
Although the exemplar is randomly chosen, it is able to improve the deraining performance, as shown in Figure~\ref{fig:The effectiveness of exemplar.}.
The number of channels is set as $32$ and the nonlinear activation we used is LeakyReLU with $\alpha = 0.2$ for all convolution layers except for the last one.
For the last layer, the channel is $3$ without any activation function.
The patch size of input is $64\times64$, and the batch size is $8$.
We use ADAM~\cite{adam} to optimize our network parameters.
The initial learning rate is 0.0005, and the rate will be divided by $10$ at $300$ and $400$ epochs, and the model training terminates after $500$ epochs.
The number of sub-network $N$ is 4 and the number of CTResBlocks $M$ in each sub-network is 8.
The patch size $l$ is 3 and the stride length $s$ is 3.
The number of nearest neighbors $k$ is set as 5.

\subsection{Results on Synthetic Datasets}
We compare our proposed network with eight state-of-the-art methods on five synthetic datasets, and the results are summarized in Table~\ref{tab: the results in synthetic datasets}.
It can be seen that our proposed method achieves the best results on all the datasets in terms of PSNR and SSIM.
Moreover, we note that the number of parameters of our model is 62\% less compared with DCSFN~\cite{mm20_wang_dcsfn}, while the SSIM is increased by 1.62\% on the Rain200H dataset.
In Figure~\ref{fig:deraining-syn-example}, we also visualize several deraining results on the synthetic datasets. It can be seen that our method is able to restore better details and textures and obtain cleaner and clearer background images, while other approaches hand down some rain streaks or lose some details.
\begin{figure*}[!t]
\begin{center}
\begin{tabular}{cccccccccc}
\includegraphics[width = 0.243\linewidth]{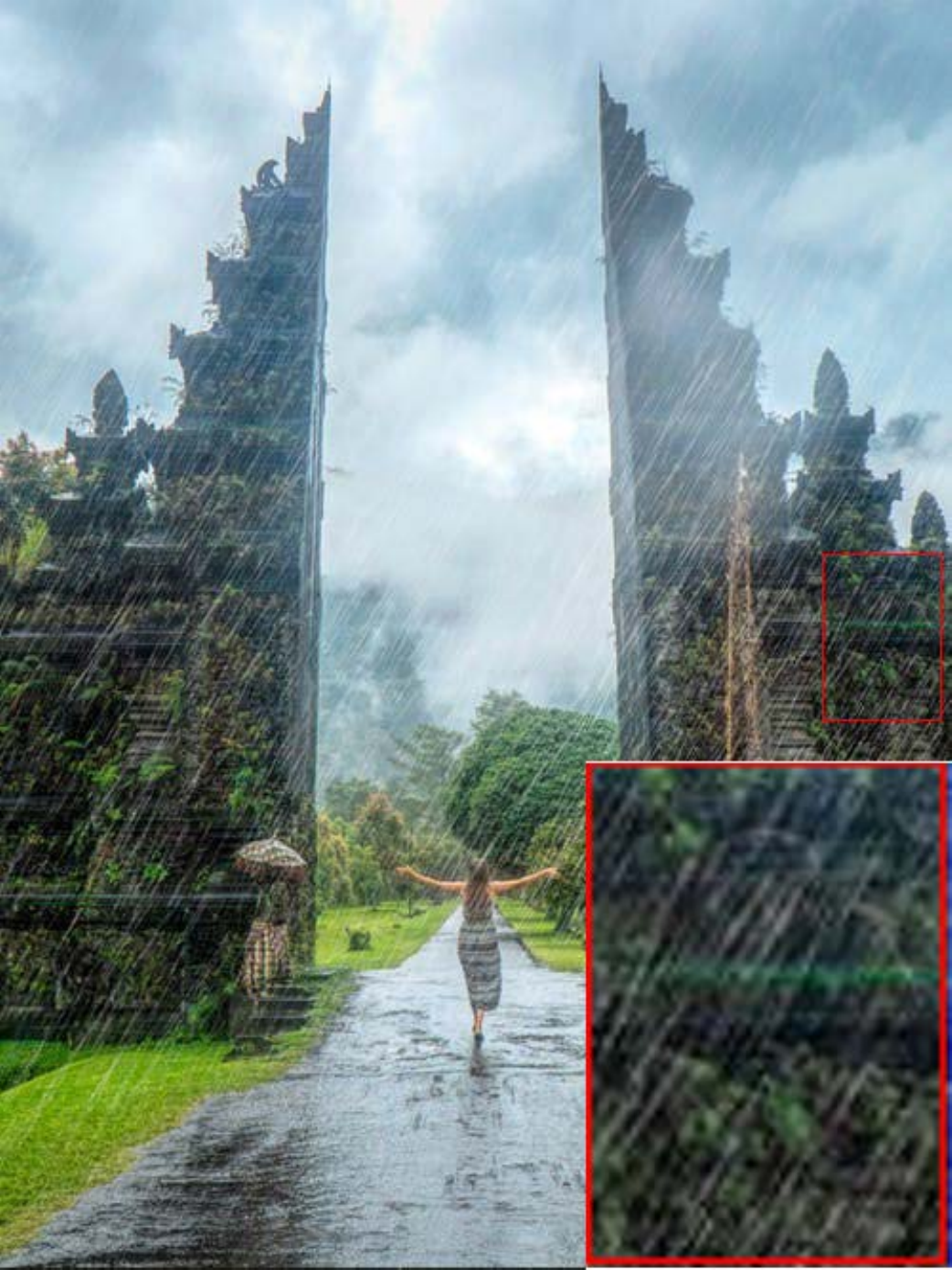} &\hspace{-4.5mm}
\includegraphics[width = 0.243\linewidth]{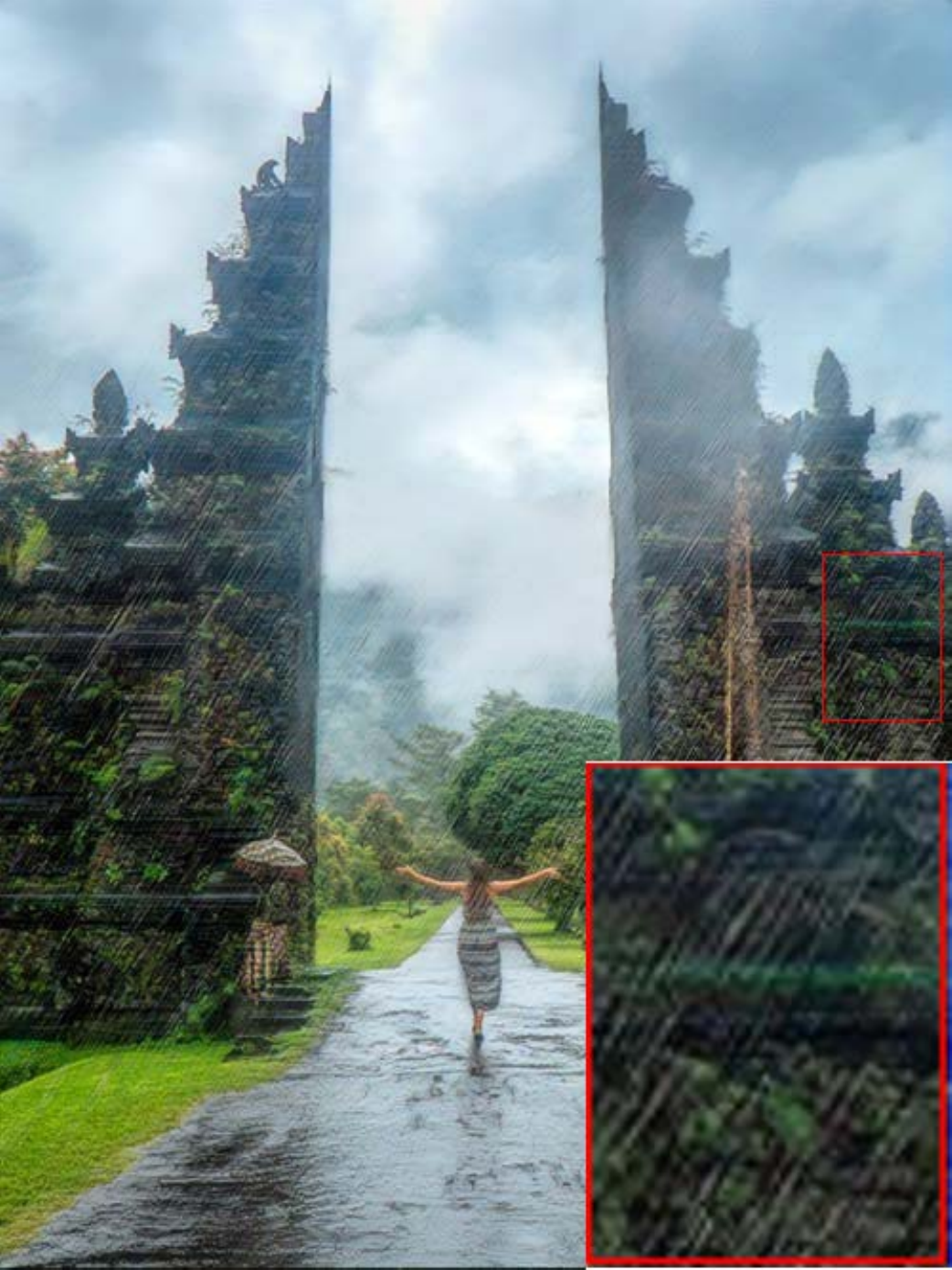} &\hspace{-4.5mm}
\includegraphics[width = 0.243\linewidth]{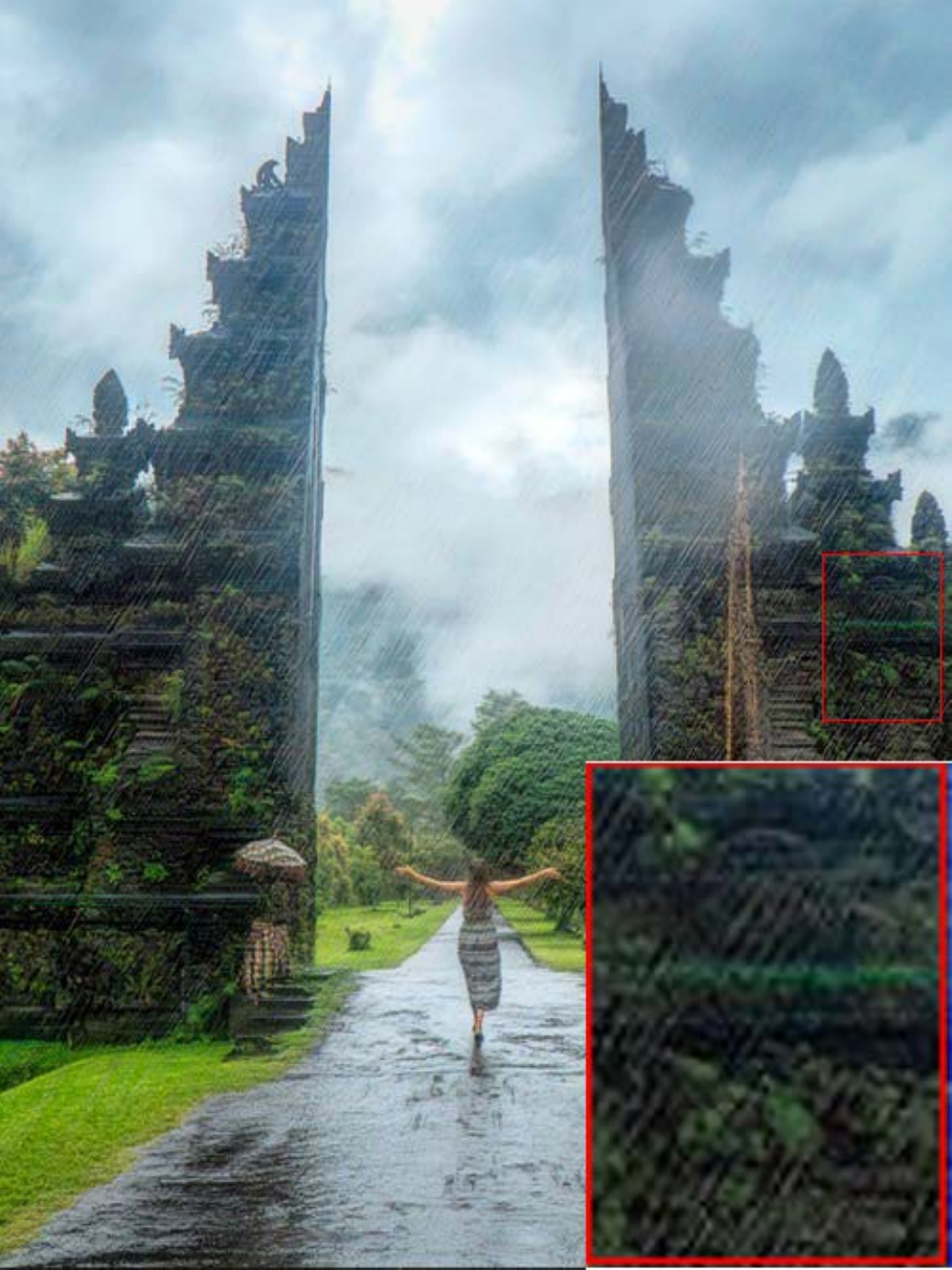} &\hspace{-4.5mm}
\includegraphics[width = 0.243\linewidth]{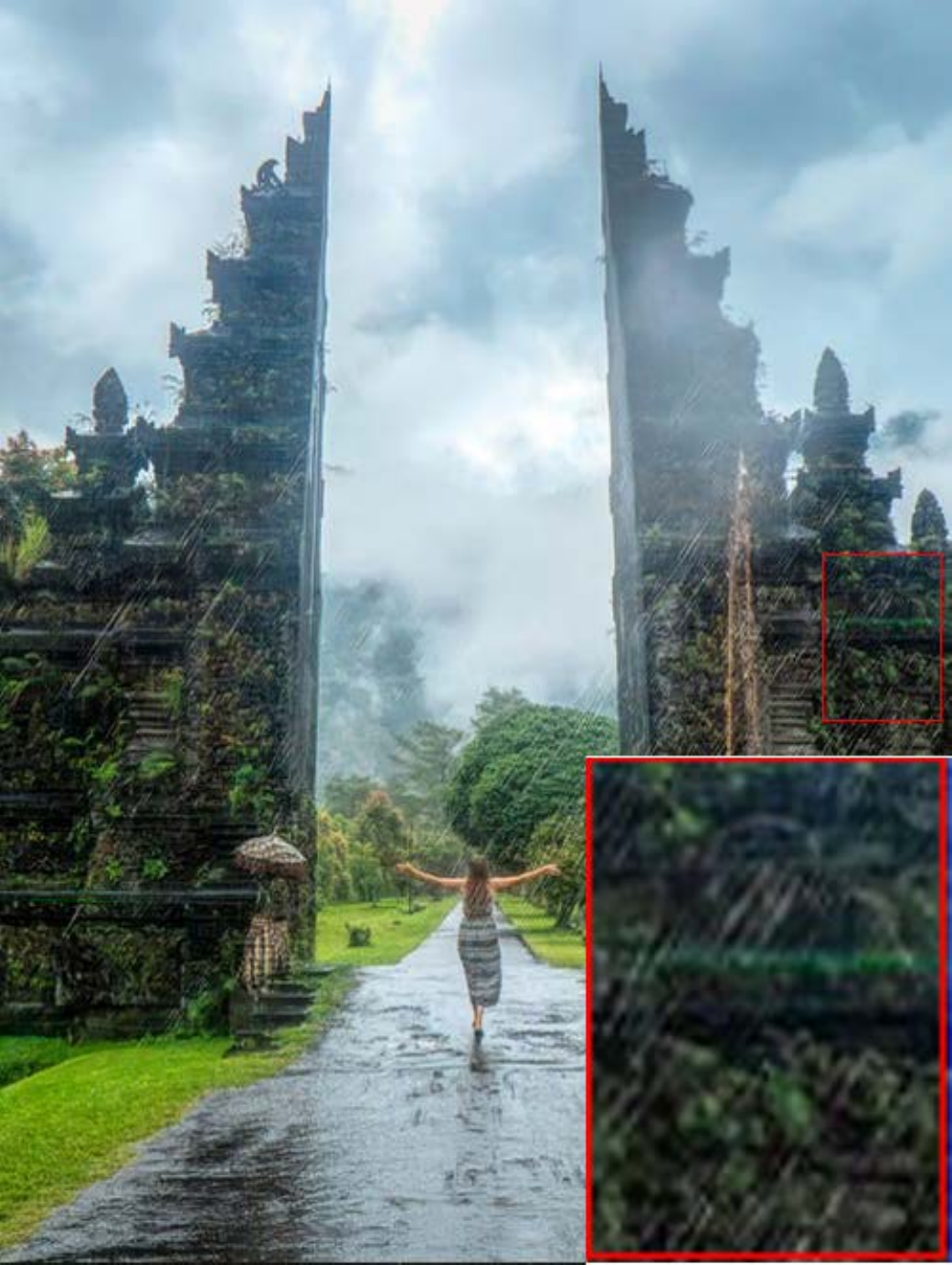} 
\\
(a) Input  &\hspace{-4.5mm} (b)  SpaNet&\hspace{-4.5mm} (c) DCSFN~ &\hspace{-4.5mm} (d) MSPFN
\\
\includegraphics[width = 0.243\linewidth]{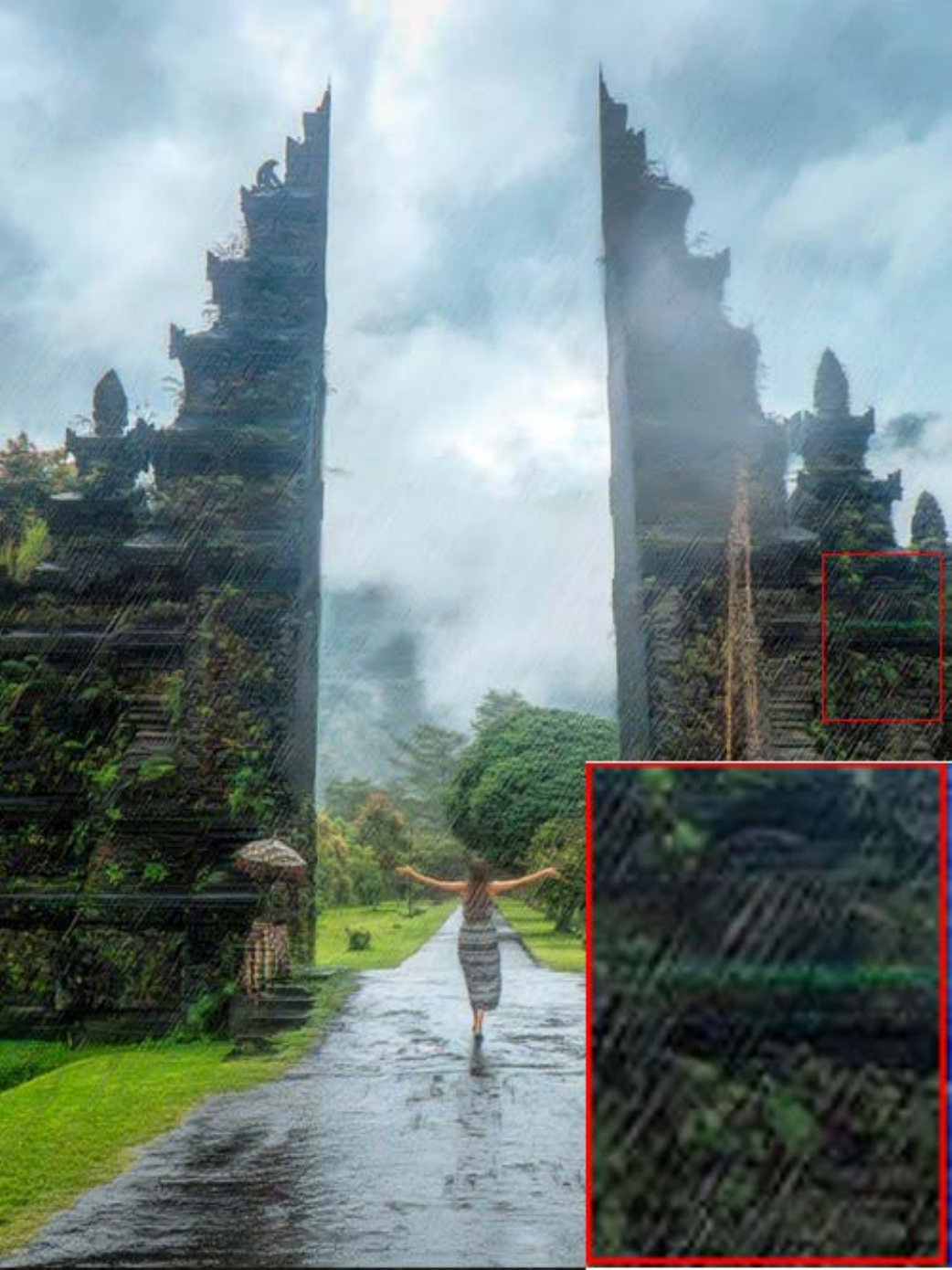} &\hspace{-4.5mm}
\includegraphics[width = 0.243\linewidth]{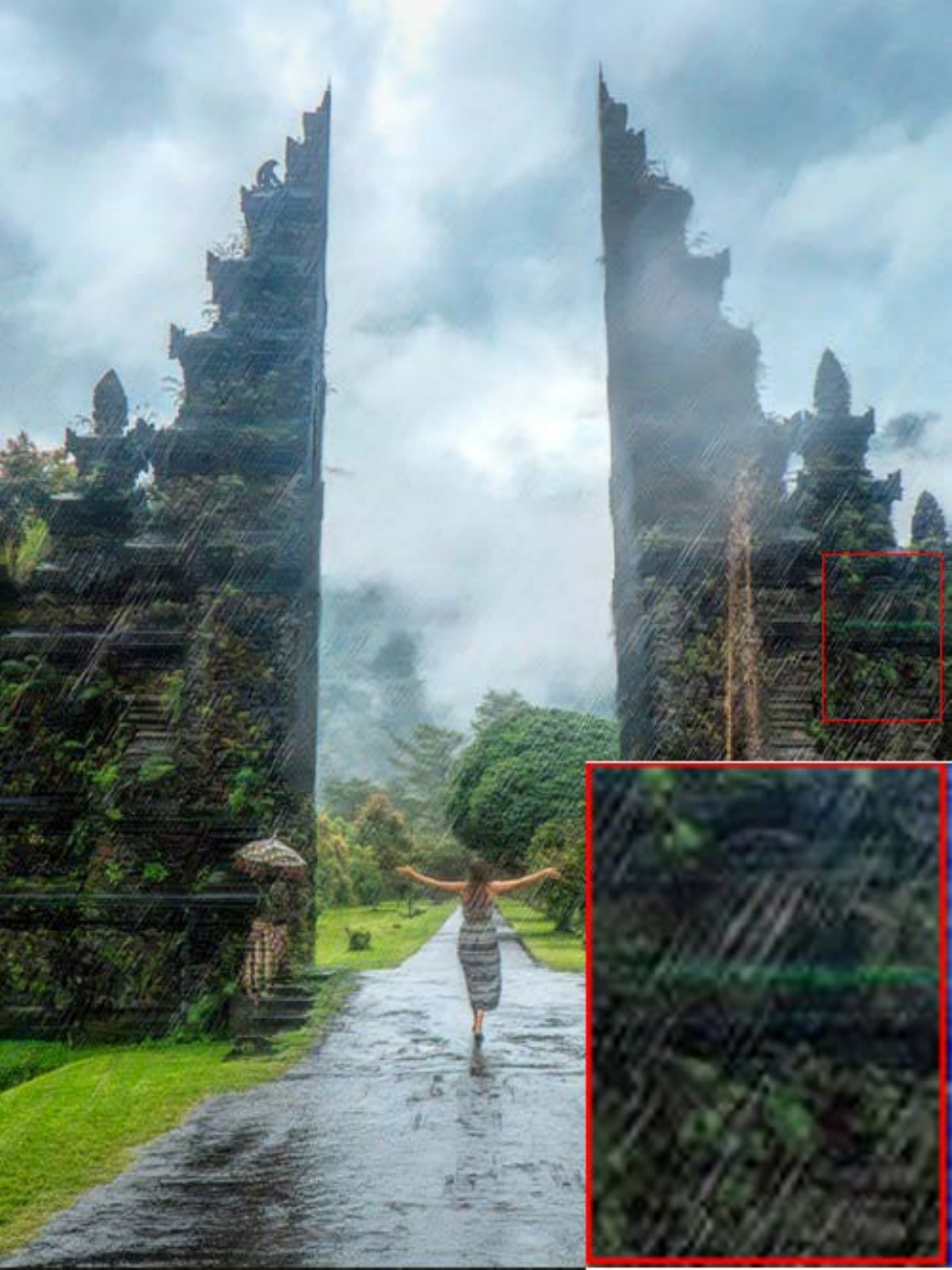} &\hspace{-4.5mm}
\includegraphics[width = 0.243\linewidth]{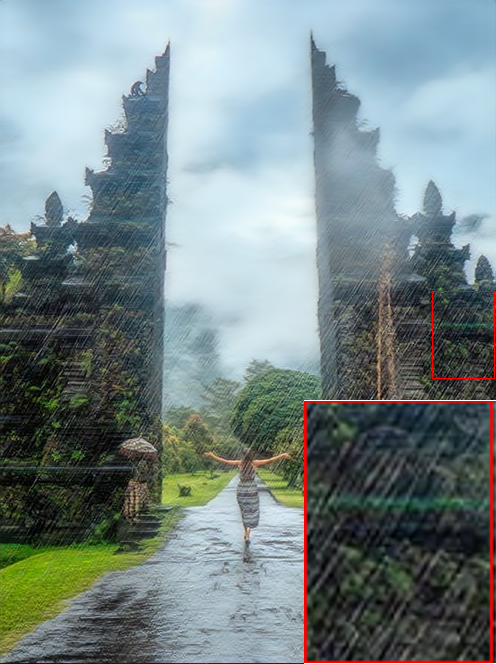} &\hspace{-4.5mm}
\includegraphics[width = 0.243\linewidth]{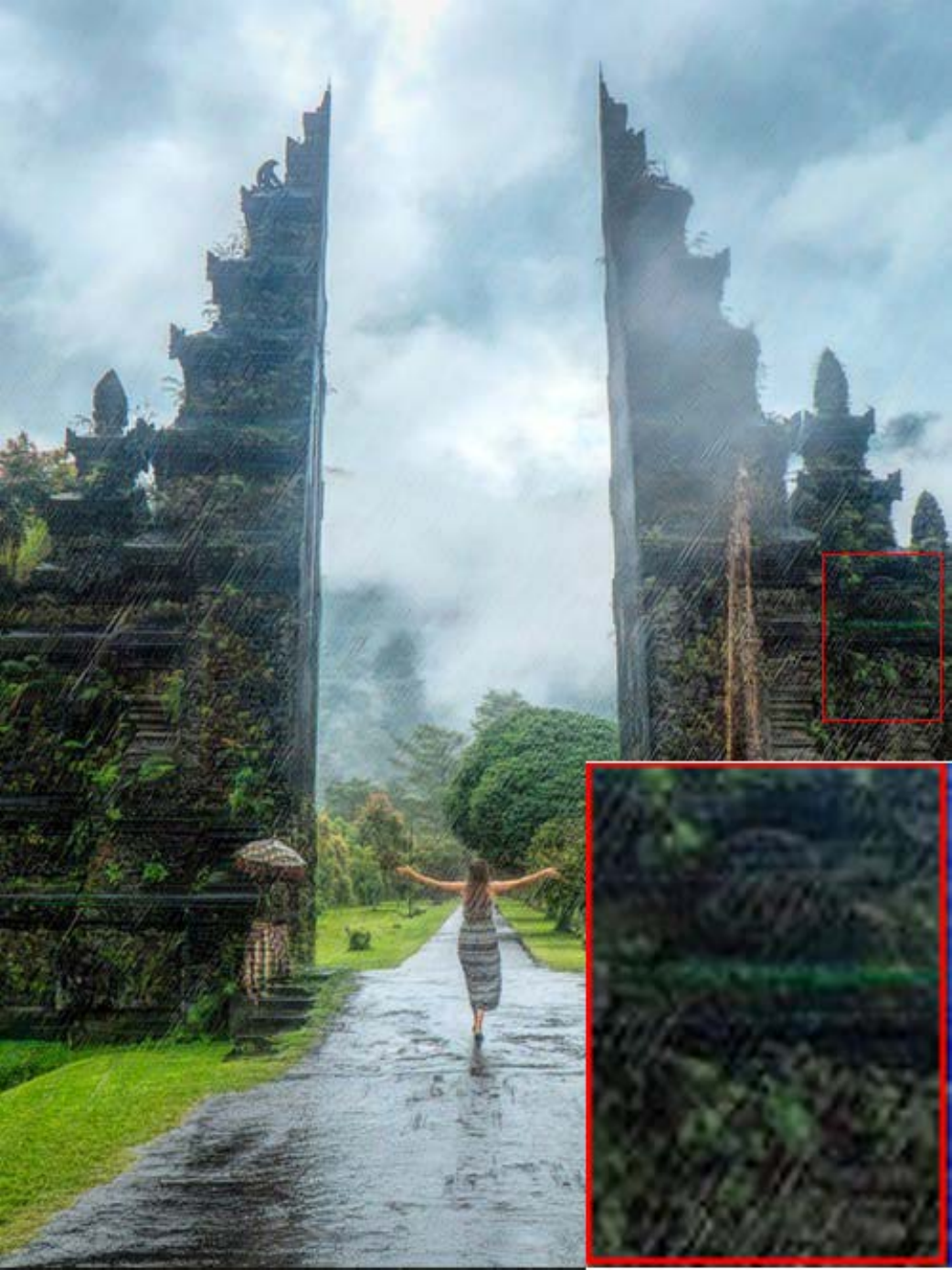} 
\\
(e) RCDNet&\hspace{-4.5mm} (f) DualGCN&\hspace{-4.5mm}(g) MOSS&\hspace{-4.5mm} (h) \textbf{MSGNN}
\end{tabular}
\end{center}
\caption{Comparisons with state-of-the-art methods on real-world dataset.
The proposed MSGNN is able to produce a clearer result.
}
\label{fig:deraining-real-example}
\end{figure*}
\subsection{Results on the Real-world Dataset}
We further demonstrate the effectiveness of the proposed method by the comparison with state-of-the-art methods on the real-world dataset in Figure~\ref{fig:deraining-real-example}.
It can be seen that the proposed algorithm is able to generate cleaner and clearer deraining results, while other methods hand down some rain streaks.
This shows the effectiveness of the proposed method in deraining real-world images.
\subsection{Ablation Study}\label{sec:Ablation Study} 

\begin{figure}[!t]
\begin{center}
\begin{tabular}{cc}
\includegraphics[width = 0.985\linewidth]{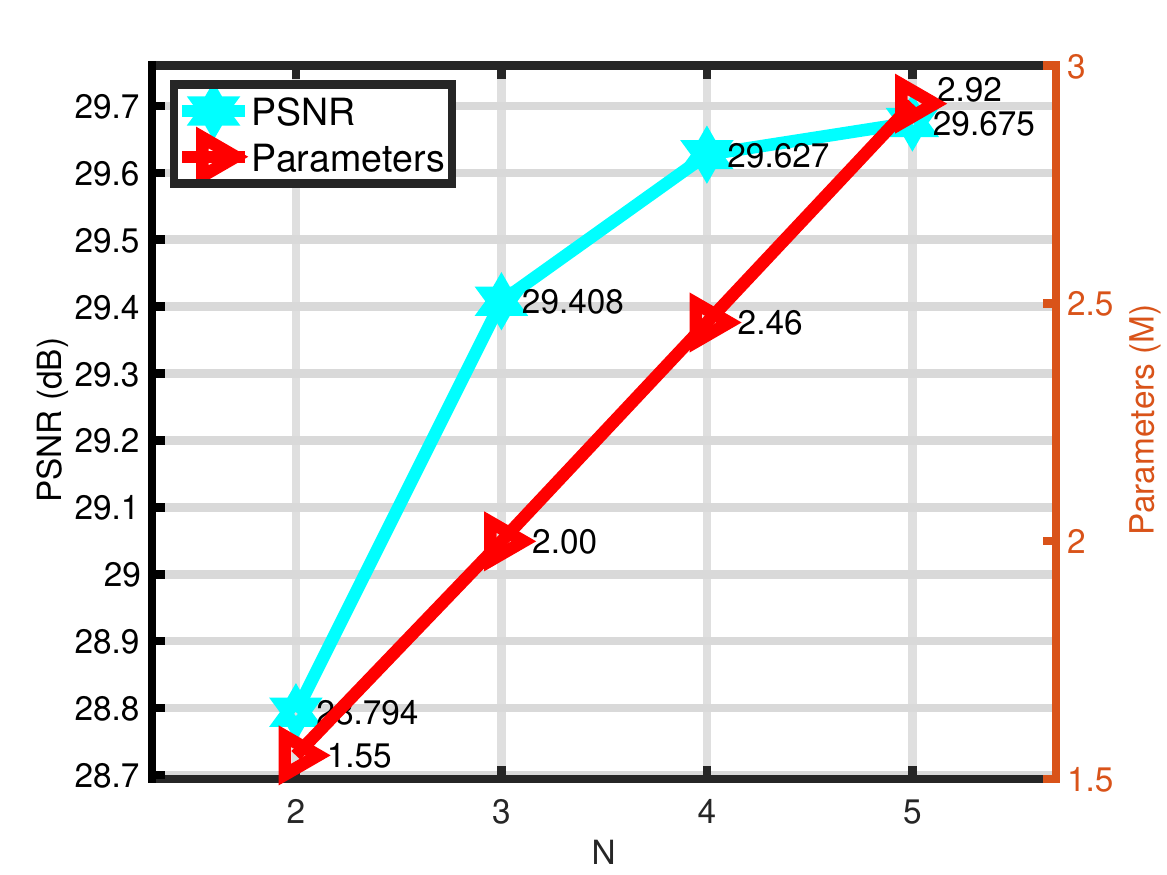}   
\end{tabular}
\end{center}
\caption{Results on the number of sub-networks.
}
\label{fig:The analysis on the number of sub-networks.}
\end{figure}

\begin{figure}[!t]
\begin{center}
\begin{tabular}{cccccc}
\includegraphics[width = 0.32\linewidth]{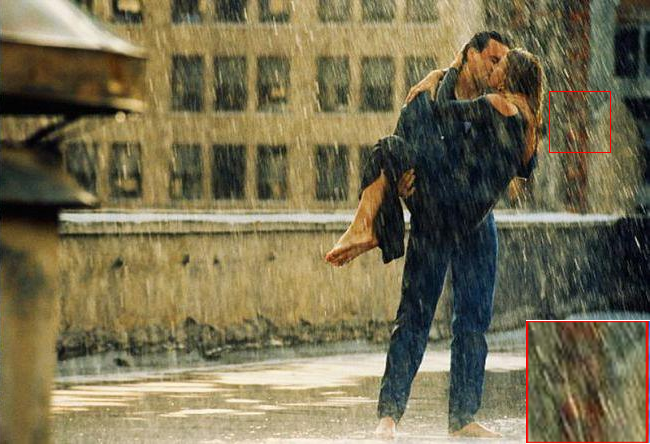} &\hspace{-4.5mm}
\includegraphics[width = 0.32\linewidth]{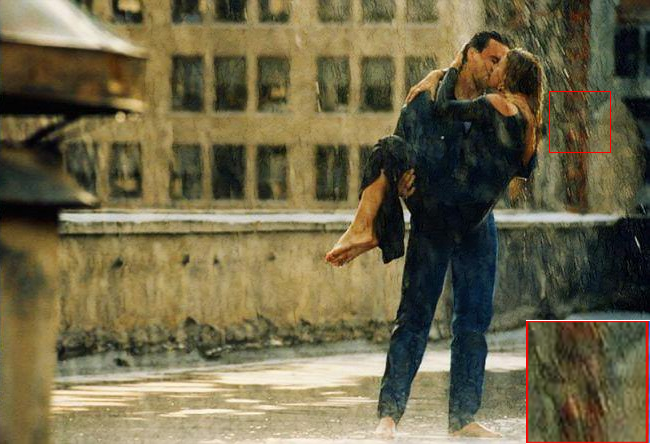} &\hspace{-4.5mm}
\includegraphics[width = 0.32\linewidth]{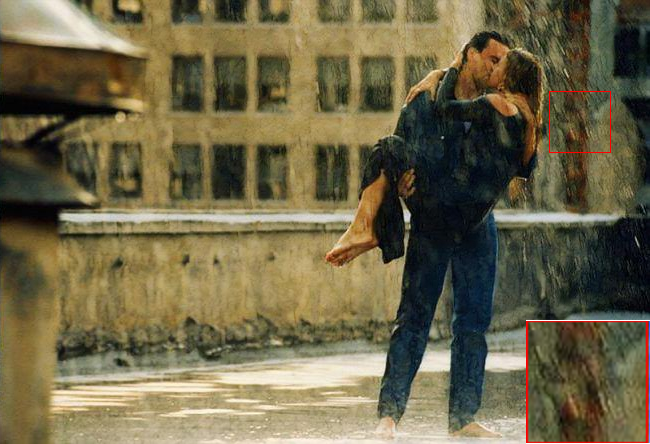} 
\\
(a) Input  &\hspace{-4.5mm} (b) w/o 1&\hspace{-4.5mm} (c) w/o $\frac{1}{2}$\&$\frac{1}{4}$ 
\vspace{1mm}
\\

\includegraphics[width = 0.32\linewidth]{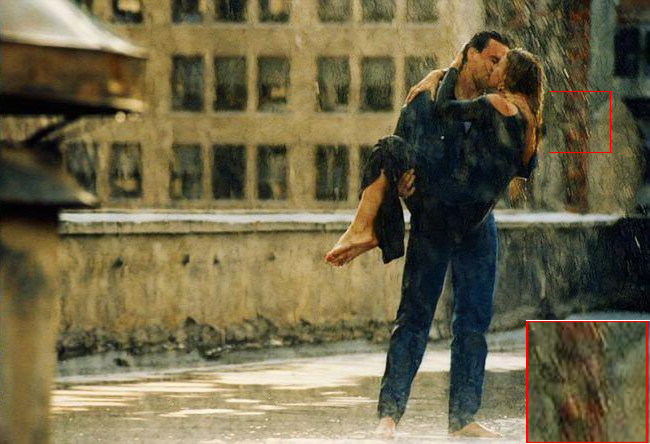} &\hspace{-4.5mm}
\includegraphics[width = 0.32\linewidth]{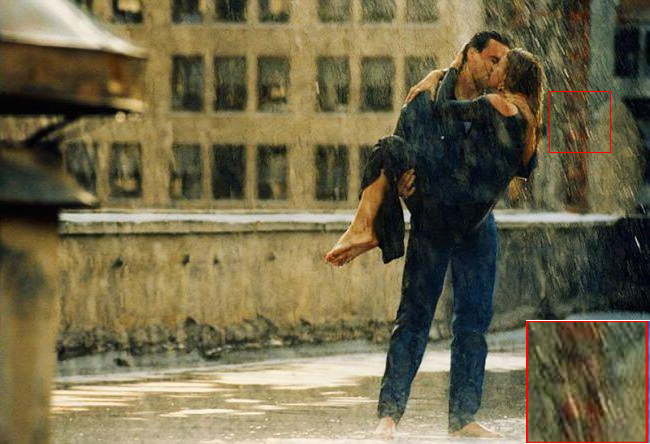} &\hspace{-4.5mm}
\includegraphics[width = 0.32\linewidth]{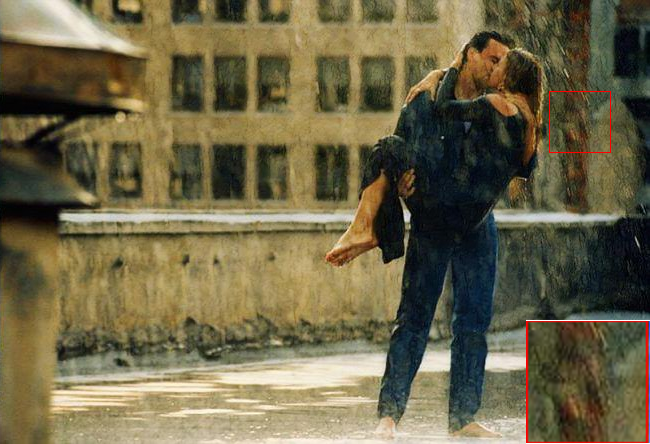} 
\\
 (d) w/o $\frac{1}{2}$ &\hspace{-4.5mm} (e)  w/o $\frac{1}{4}$&\hspace{-4.5mm} (f) Ours
\end{tabular}
\end{center}
\caption{The effectiveness of exploiting multi-scale images on a real-world image.
The proposed internal multi-scale image strategy is able to help better deraining.
}
\label{fig:The effectiveness of multi-scale.}
\end{figure}

\noindent {\bf Analysis on the Basic Components.}
In our model, we design the graph-based deep network using the fusion connection,  channel attention (Channel Transformation), and graph model.
We analyze them and present the results in Table~\ref{tab:The analysis on basic component of the network.}.
First, it can be seen that the results between $M_{1}$ and $M_{2}$ show that the fusion connection (FC) can improve the deraining performance.
Moreover, as a comparison, we also analyze other channel attention modules, ECA~\cite{cvpr_wang_eca} and SE~\cite{se}.
We can see that ECA is less effective for our model and SE can improve the deraining results in terms of PSNR and SSIM.
Furthermore, CT is the best channel attention module among the three.
At last, we can observe that our graph model can boost the deraining results.
Hence, the fusion connection, CT, and graph modules applied in our framework are all meaningful for deraining.
\begin{table}[!t]
\label{tab:The analysis on basic component of the network.}
\scalebox{0.81}{
\centering
		\begin{tabular}{lcccccc}
			\toprule
			Experiments  &$M_{1}$ & $M_{2}$ & $M_{3}$     &$M_{4}$ &$M_{5}$&$M_{6}$            \\
			\midrule
			FC         &  \XSolidBrush      & \CheckmarkBold & \CheckmarkBold & \CheckmarkBold &\CheckmarkBold  & \CheckmarkBold                                \\
			ECA       &  \XSolidBrush      &     \XSolidBrush         & \CheckmarkBold&\XSolidBrush              &\XSolidBrush &\XSolidBrush       \\
			SE         & \XSolidBrush       &  \XSolidBrush            & \XSolidBrush            &  \CheckmarkBold&\XSolidBrush& \XSolidBrush                          \\
			CT         &\XSolidBrush        & \XSolidBrush             & \XSolidBrush            &  \XSolidBrush              &\CheckmarkBold&\CheckmarkBold \\
			Graph      & \XSolidBrush       &\XSolidBrush              & \XSolidBrush            & \XSolidBrush                 &\XSolidBrush&       \CheckmarkBold                    \\
			\midrule
			PSNR        & 29.161  & 29.451       & 27.535     & 29.397      & 29.488 &   29.627   \\
			SSIM       & 0.9111   & 0.9147       & 0.8944     & 0.9149    &  0.9163   &  0.9178  \\
			\bottomrule
		\end{tabular}
	}
 \caption{Ablation results on different components.
The \CheckmarkBold and \XSolidBrush symbols denote the corresponding component whether is adopted or not.}
\end{table}
\\
{\bf Effect of the Stride Length.}
In Table~\ref{tab:The effect of stride.}, we present the results of different strides.
Since the number of graph patches with stride 1 is increased sharply when the training patch size is $64 \times 64$, which is out of our computer memory,
we adjust the training patch size to $48 \times 48$.
We can observe that the model achieves the best results when the stride is 3 under the two conditions.
The stride being 3 means that the obtained graph patches are not overlapped because the graph patch size is also 3.
Moreover, we find an interesting phenomenon from Table~\ref{tab:The effect of stride.} that the deraining results are influenced by the training patch size, and the larger the training patch size, the better the results. 
\begin{table}[!t]
\centering
\scalebox{0.9}{
\begin{tabular}{cccc|ccc}
\toprule
  &\multicolumn{3}{c}{Training Patch is $64 \times 64$} &\multicolumn{3}{c}{Training Patch is $48 \times 48$} \\
\midrule
 $s$  & 1& 2& 3 &  1& 2& 3  \\
\midrule
PSNR  & -& 29.472&  29.627  &   29.011& 28.942& 29.016 \\
SSIM  &-&  0.9168&  0.9178   & 0.9116&  0.9112& 0.9119\\
\bottomrule
\end{tabular}}
\caption{Effect of the stride length.
}
\label{tab:The effect of stride.}
\end{table}
\\
{\bf Effect of the Patch Size.}
We analyze the effect of the size of the graph patch, i.e., $l$, and the results are reported in Table~\ref{tab:The effect of the size of patch of graph.}.
We can see the deraining performance reaches its best when the patch size is 3 in both PSNR and SSIM compared to the cases of patch size being 5 and 7.
Further, we can observe that the PSNR decreases as $l$ increases.
This demonstrates that a bigger patch size cannot help improve the deraining performance.
As such, we set the default graph patch size as 3.
\begin{table}[!t]
\centering

\setlength{\tabcolsep}{6mm}
\scalebox{0.99}{
\begin{tabular}{ccccccc}
\toprule
$l$  & $3$& $5$& $7$   \\
\midrule
PSNR  &  29.627& 29.606& 29.584    \\
SSIM  & 0.9178&  0.9175& 0.9178   \\
\bottomrule
\end{tabular}}
\caption{Effect of the size of graph patch.
}
\label{tab:The effect of the size of patch of graph.}
\end{table}
\\
{\bf Effect of the Number of Nearest Neighbors.}
\begin{table}[!t]
\centering

\setlength{\tabcolsep}{6mm}
\scalebox{0.99}{
\begin{tabular}{ccccccc}
\toprule
$k$  & $3$& $5$& $7$    \\
\midrule
   PSNR & 29.501&  29.627&29.565   \\
SSIM  &0.9167&   0.9178&0.9172   \\
\bottomrule
\end{tabular}}
\caption{Analysis on the number of nearest neighbors.
}
\label{tab:The analysis on the number of KNN.}
\end{table}
We analyze the effect of the number of nearest neighbors, i.e., $k$, and the results are presented in Table~\ref{tab:The analysis on the number of KNN.}.
It can be observed that the results are the best when $k=5$ in terms of PSNR and SSIM.
And, both $k=3$ and $k=7$ perform worse than $k=5$.
Hence, we select $k=5$ as the default setting of the network.
\\
{\bf Analysis of the Number of Sub-networks.}
We analyze the effect of the number of sub-networks $N$, and the results are presented in Figure~\ref{fig:The analysis on the number of sub-networks.}.
It can be seen that $N=4$ achieves good results and the deraining performance will not be significantly improved as $N$ increases in terms of PSNR. Therefore, we set $N=4$ as our default setting due to the trade-off between the deraining performance and the number of parameters.
Note that the model also achieves state-of-the-art performance compared with other methods on the most challenging Rain200H~\cite{derain_jorder_yang} when $N=2$, while the number of parameters only has 1.55M.
\\
{\bf Analysis on the Use of Multi-Scale Images.}
Since our graph neural network is based on the multi-scale aggregation module, we provide an analysis on its effectiveness.
We show the results on the real-world dataset in Figure~\ref{fig:The effectiveness of multi-scale.}.
We can observe that the default model with all scales obtains clearer and cleaner results, while other models always hand down some rain streaks.
Hence, the use of multi-scale input images in our framework is helpful for the deraining task.

\begin{figure}[!t]
\begin{center}
\begin{tabular}{ccc}
\includegraphics[width = 0.32\linewidth]{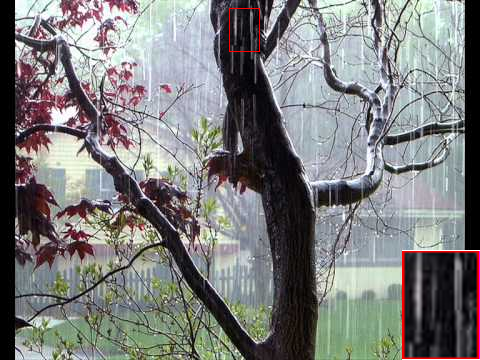} &\hspace{-4.5mm}
\includegraphics[width = 0.32\linewidth]{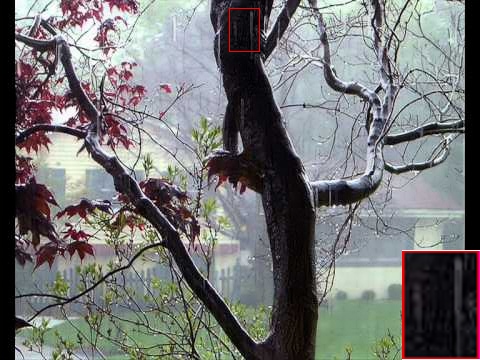} &\hspace{-4.5mm}
\includegraphics[width = 0.32\linewidth]{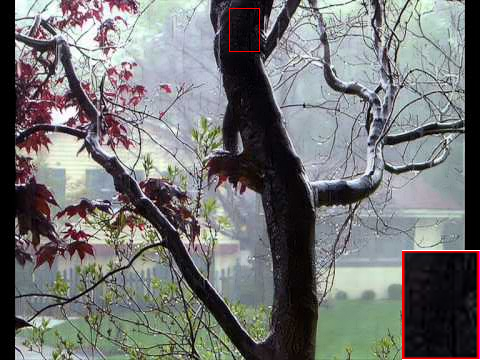} 
\\
\includegraphics[width = 0.32\linewidth]{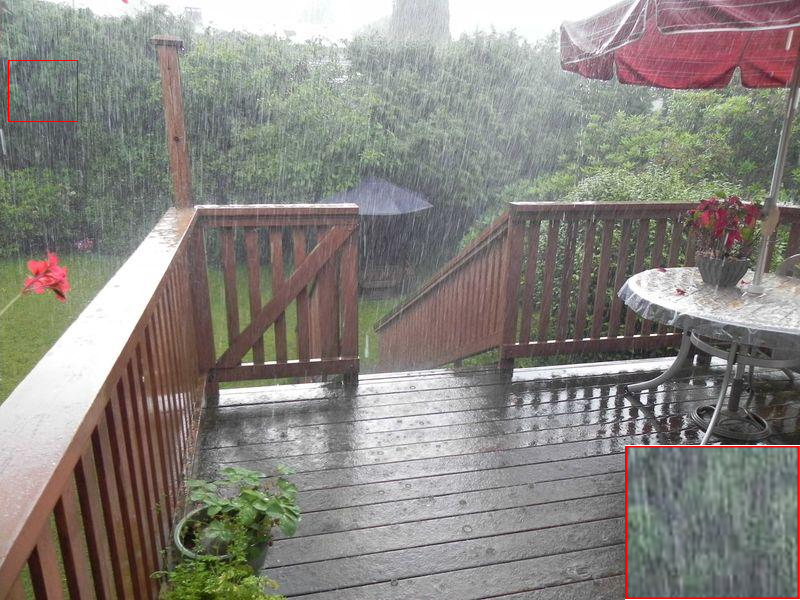} &\hspace{-4.5mm}
\includegraphics[width = 0.32\linewidth]{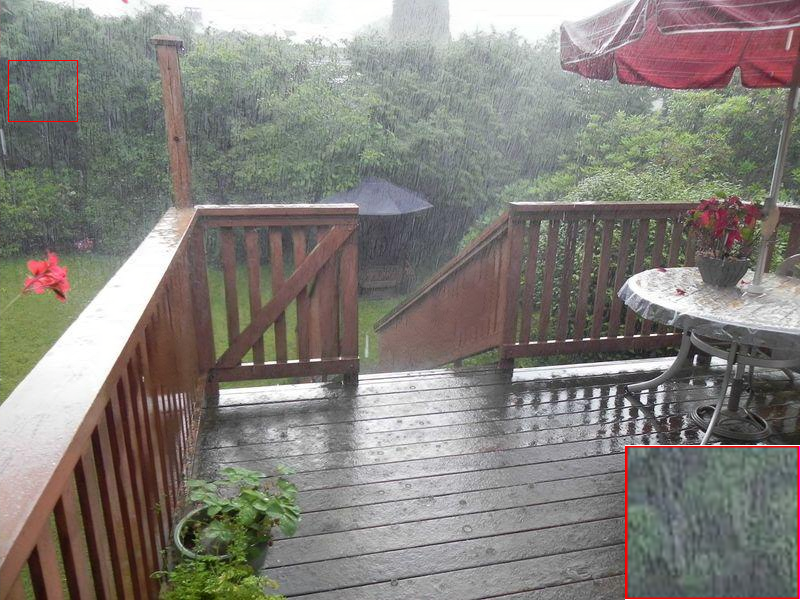} &\hspace{-4.5mm}
\includegraphics[width = 0.32\linewidth]{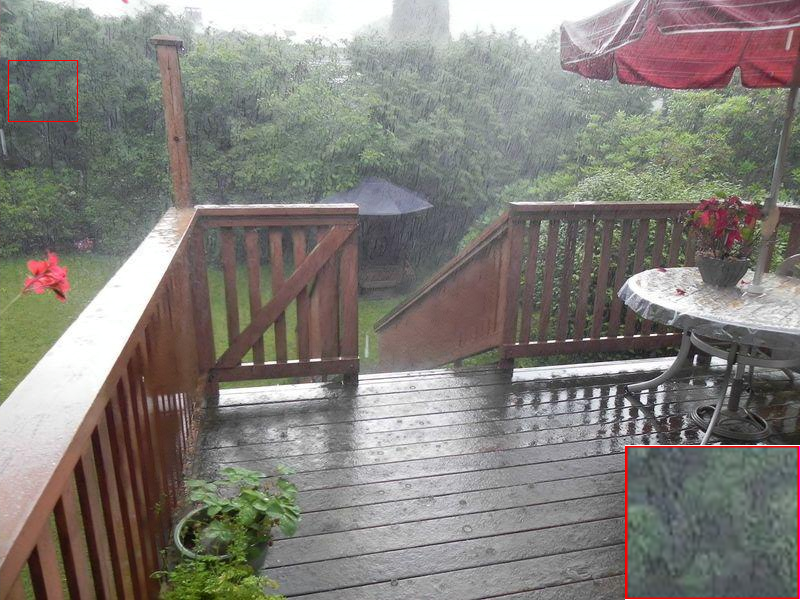} 
\\
(a) Input &\hspace{-4.5mm} (b) w/o exemplar&\hspace{-4.5mm} (c) w/ exemplar 
\end{tabular}
\end{center}
\caption{Effectiveness of the exemplar on the real-world dataset.
The proposed external images as exemplars are able to improve image deraining quality.
}
\label{fig:The effectiveness of exemplar.}
\end{figure}

\noindent {\bf Effect of the Exemplar.}
Here, we demonstrate the effectiveness of the exemplar on deraining, and the results are shown in Figure~\ref{fig:The effectiveness of exemplar.}.
We can observe that the model with an exemplar is able to improve the deraining performance, while the model without an exemplar hands down some rain streaks.
It illustrates that our proposed model with exemplars can effectively utilize more information on rainy images, which makes the model more robust under different rainy conditions.

\section{Conclusion}
In this paper, we have proposed a deep multi-scale graph neural network with exemplars, called MSGNN, for image deraining.
By exploiting internal non-local similarity in multi-scale input images and external non-local similarity in an exemplar image with a graph model and attention mechanism, our model achieves favorable results on both synthetic and real-world datasets compared with state-of-the-art methods, demonstrating the effectiveness of our deraining model. 
In future work, we plan to explore different strategies for selecting the exemplar image for better deraining.

\section*{Acknowledgements}
This work was supported by the National Natural Science Foundation of China No.62306343, Shenzhen Science and Technology Program (No.KQTD20221101093559018).

\bibliographystyle{named}
\bibliography{ijcai24}

\end{document}